\documentclass[11pt]{article}

\usepackage[preprint]{acl}

\usepackage{times}
\usepackage{latexsym}
\usepackage[T1]{fontenc}
\usepackage[utf8]{inputenc}
\usepackage{microtype}
\usepackage{inconsolata}
\usepackage{graphicx}
\usepackage{booktabs}
\usepackage{amsmath,amssymb}
\usepackage{xcolor}
\usepackage{colortbl}
\usepackage{tabularx}
\renewcommand{\dh}[1]{\textcolor{black}{#1}}
\usepackage{multirow}
\usepackage{algorithm}
\usepackage{algpseudocode}
\usepackage{tikz}
\usetikzlibrary{arrows.meta,positioning,calc,patterns}
\usepackage{pgfplots}
\pgfplotsset{compat=1.18}
\usepgfplotslibrary{groupplots}
\usepackage{placeins}

\newif\ifoutline
\outlinetrue
\ifdefined\SUBMIT \outlinefalse \fi
\ifoutline
  \usepackage{kotex}
  \newcommand{\outlinenote}[1]{{\small\color{blue!55!black}#1\par}}
  \newcommand{\checknote}[1]{{\small\color{red!75!black}\textbf{[Check]} #1\par}}
\else
  \newcommand{\outlinenote}[1]{}
  \newcommand{\checknote}[1]{}
\fi

\newcommand{\KL}{\mathrm{KL}}

\title{N-Gram Memories at Two Bits: Quantizing Lookup Tables in Large Language Models}

\author{
  Donghoon Han$^{1}$ \quad
  Minseop Jung$^{1}$ \quad Dong Woo Kwak$^{1}$ \quad SeungJae Lee$^{1}$ \quad Junhoo Lee$^{2}$\thanks{Corresponding author.}  \\
  $^{1}$Dnotitia Inc. \quad $^{2}$KAIST \\
  \texttt{\{dhk1349,\,msjung,\,HelloDongWoo,\,seungjae.lee\}@dnotitia.com} \\
  \texttt{junhoo.lee@kaist.ac.kr}
}

\begin{document}
\maketitle

\begin{abstract}
Conditional memory expands language-model capacity through selective lookup, but the full embedding table must remain available for every possible input.
We introduce Conditional Program (CP), a compressed representation that reconstructs only the rows an input requests.
CP represents each row by a short code and a local scale, with a reconstruction program shared across the table.
We instantiate this representation with a Gaussian trellis, norm restoration and independently decodable row records.
CP's two-bit code lowers output KL by 41--63\% relative to 2-bit round-to-nearest.
Analysis across three trained n-gram memory families shows that Gaussian-like coordinate statistics coexist with learned relationships among rows; at matched storage budgets, the shared trellis outperforms the fitted codebooks and low-rank residual variants we test.
For DeepSeek, it reduces table storage from 203\,GB to 51\,GB, leaving room for larger workloads when the table is kept on GPU.
\end{abstract}

\section{Introduction}
\label{sec:intro}

Conditional memory expands language-model capacity without a proportional increase in per-token computation.
It retrieves learned representations from large embedding tables, allowing a model to use only a small part of its memory for each input \citep{cheng2026engram,liu2026scalingembeddings}.
This makes higher capacity attractive under limited compute budgets, but the tables still occupy substantial storage.
Moving them from accelerator memory to host memory changes where that cost is paid without reducing it.

Compressing conditional memory presents a different challenge from conventional LLM weight compression.
Although its embedding tables contain learned parameters, they are accessed through sparse, input-dependent lookups rather than dense computation.
Existing LLM compression methods primarily target dense model weights, leaving the distinct access patterns of conditional memory largely unaddressed.
Conditional memory instead requires reducing the footprint of the entire table while preserving independent access to individual rows and the information they provide to the model.
Can conditional memory be stored more compactly while reconstructing only the entries needed by the current input?

We propose \textbf{Conditional Program (CP)}, a representation that stores instructions for reconstructing each entry through a shared decoder.
Our hypothesis is that a compact shared reconstruction rule, together with short row-specific codes and local scales, can preserve the signal that the model consumes from the table.
An input selects which entries to reconstruct, so the extra work is confined to the rows it accesses.
This exchanges storage across the full table for reconstruction work on a small subset, carrying the selective use of conditional memory into its compressed representation.

\begin{figure*}[t]
  \centering
  \includegraphics[width=\textwidth]{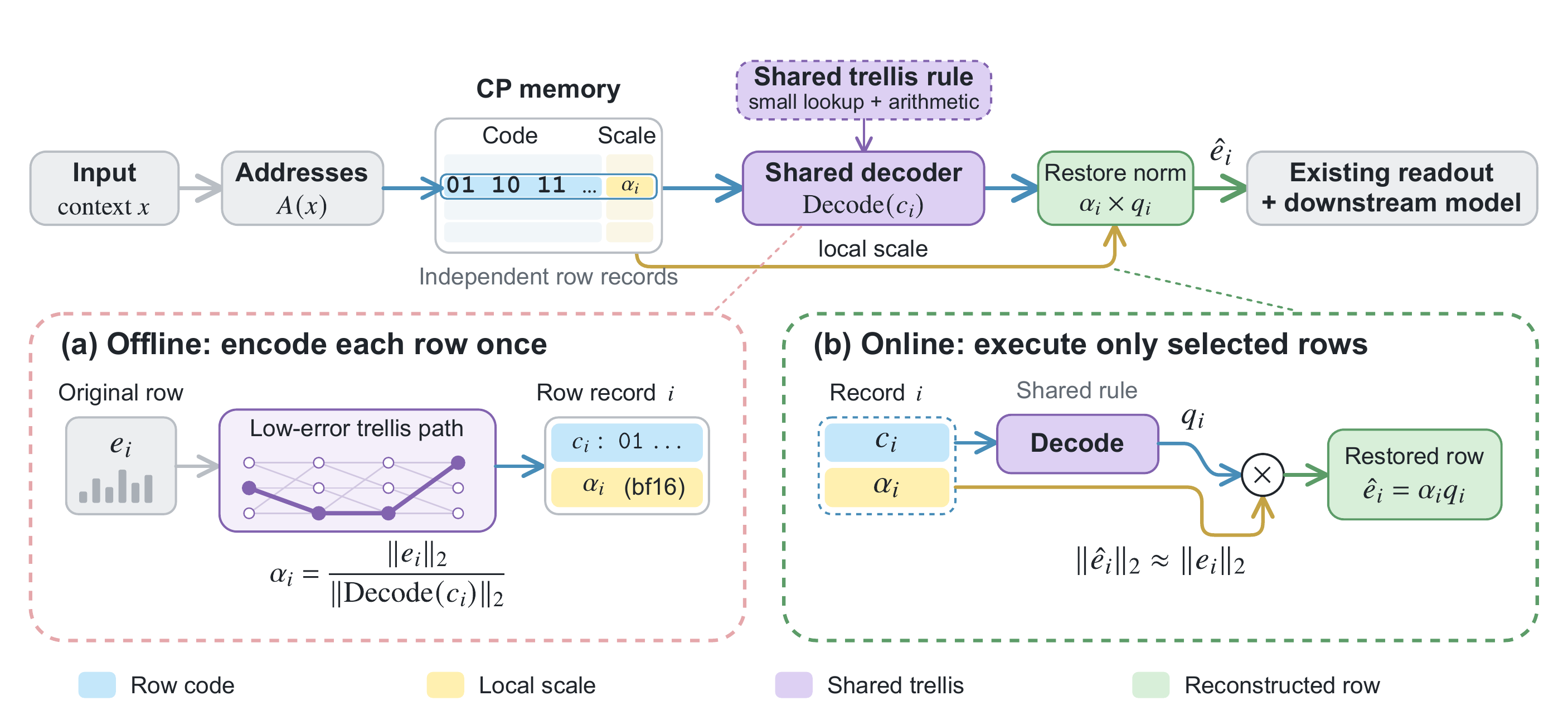}
  \caption{\textbf{Conditional Program (CP).} An input selects row addresses in the compressed memory. Each record contains a trellis code and a bf16 scale; a shared decoder reconstructs the vector, and the scale restores its norm before the existing readout. \textbf{(a)} Offline, the encoder selects a low-error trellis path and stores its code with the norm-restoring scale. \textbf{(b)} Online, only addressed records are fetched, decoded and scaled. Trellis states and paths are schematic.}
  \label{fig:cp-overview}
\end{figure*}

Specifically, CP represents each row with a short code and a local scale, using a decoder shared across the table.
We model normalized row contents as the accumulation of many small learning contributions dispersed across coordinates; when these contributions are weakly dependent and none dominates, a Gaussian source approximation follows from a central-limit argument.
Therefore, we encode rows in their stored coordinate system with a shared Gaussian trellis \citep{tseng2024qtip}.
The local scale restores the decoded vector's norm, and independently stored row records allow the runtime to fetch and reconstruct only requested entries (Figure~\ref{fig:cp-overview}).
\dh{At two bits, CP reduces table storage by roughly four- to eightfold, with most reported task scores remaining close to their original values.}
\dh{For a 4,096-token DeepSeek prefill from host memory, its row layout reduces lookup time from 10.78 to 6.54\,ms compared with a QTIP tile-window decoder in a layout microbenchmark.}

Our contributions are as follows:
\begin{itemize}
  \item We introduce Conditional Program, a representation that exchanges storage across the full memory table for reconstruction work on the few entries selected by each input.
  \item We instantiate CP with a shared Gaussian trellis, per-row norm restoration and independently addressable records, preserving selective access in the compressed representation.
  \item \dh{We evaluate downstream performance across diverse tasks, analyze compression fidelity, and measure lookup costs and workload capacity under GPU and host-memory placement.}
\end{itemize}

\section{Preliminaries}
\label{sec:background}

\subsection{Conditional Memory as Selective Lookup}
\label{sec:bg-memories}

We consider conditional-memory table $E \in \mathbb{R}^{N \times d}$, where each row $e_i \in \mathbb{R}^d$ stores a learned representation.
For an input $x$, let $A(x) \subseteq \{1,\ldots,N\}$ denote the requested row addresses.
N-gram-based conditional memory determines row addresses by hashing input n-grams independently for each order and head. Each token retrieves a fixed number of rows, so the accessed memory is
\begin{equation}
    E_{A(x)} = \{e_i \mid i \in A(x)\},
    \qquad |A(x)| \ll N.
\end{equation}
Because the addresses are computed directly from the input tokens, retrieval does not require scanning or computing over the entire table.

The retrieved embeddings are incorporated into the model's hidden states through a readout operation.
Let $e(x)$ denote the concatenation of the retrieved rows and $h$ the current hidden state.
The readout can be expressed as
\begin{equation}
    h' = h + \mathcal{R}(h,e(x)),
\end{equation}
where $\mathcal{R}$ is a model-specific operation that may involve linear projections or input-dependent gating.
The resulting hidden state is passed to subsequent network layers, so errors in the retrieved embeddings can propagate through the readout and affect the model's final predictions.

\subsection{Compressing Conditional Memory}
\label{sec:bg-compression}

The storage footprint of a compressed table and the cost of accessing an individual row are distinct quantities.
Compression can reduce storage by encoding or transforming multiple rows jointly, but such representations may couple them within a larger reconstruction unit.
Consequently, accessing a single row may require fetching or decoding a larger block of data. Thus, a smaller compressed table does not necessarily imply efficient access under sparse lookup.

Let $\mathcal{C}(E)$ denote a compressed representation of the table and $\mathcal{D}(\mathcal{C}(E),i)$ the reconstruction of row $i$.
Preserving selective access requires that each requested row can be reconstructed without decoding unrelated rows:
\begin{equation}
    \hat{e}_i = \mathcal{D}(\mathcal{C}(E),i),
    \qquad i \in A(x).
\end{equation}
The compressed representation must therefore reduce the full table's footprint while confining data fetching and reconstruction to the requested entries.

Beyond selective access, compression must preserve the information that retrieved embeddings provide to the model.
A conventional measure of reconstruction quality is the mean squared error
\begin{equation}
    D_{\mathrm{rec}}
    =
    \frac{1}{N}
    \sum_{i=1}^{N}
    \|e_i-\hat{e}_i\|_2^2.
\end{equation}
However, reconstruction errors of similar magnitude need not produce similar changes in model predictions, since the retrieved embeddings pass through the readout and subsequent network layers.
In particular, errors in embedding magnitude and direction can have different downstream effects, as examined in Section~\ref{sec:propagation}.
Numerical reconstruction accuracy alone therefore does not fully characterize the quality of compressed conditional memory.

Let $p_E(\cdot\mid x)$ and $p_{\mathcal{C}}(\cdot\mid x)$ denote the output distributions of the original and compressed models, respectively.
We formulate conditional-memory compression as minimizing expected output divergence under a storage budget $B$:
\begin{equation}
\begin{aligned}
    \min_{\mathcal{C}} \quad
    & \mathbb{E}_{x}\left[
        D_{\mathrm{KL}}\left(
            p_E(\cdot\mid x)
            \parallel
            p_{\mathcal{C}}(\cdot\mid x)
        \right)
    \right] \\
    \text{where} \quad
    & S(\mathcal{C}) \leq B.
\end{aligned}
\end{equation}
Here, $S(\mathcal{C})$ includes compressed row codes, per-row metadata, shared reconstruction information, and storage overhead.
Each requested row must be independently reconstructible, while the address function and all remaining model parameters remain fixed.

\section{Conditional Programs}
\label{sec:method}
\label{sec:recipe}

\subsection{Conditional Program Representation}
\label{sec:method-program}
\label{sec:shared-program}

We propose Conditional Program (CP), a compressed representation that preserves the selective access pattern of conditional memory.
Rather than storing each embedding directly, CP represents each row with a short code and a reconstruction rule shared across the table.
Specifically, each row $e_i$ is represented by a code $c_i$ and a local scale $\alpha_i$, and reconstructed through a shared function $f$:
\begin{equation}
    \hat e_i = \alpha_i f(c_i).
    \label{eq:cp-reconstruction}
\end{equation}
The reconstruction function is shared across the table, while each row retains only its code and scale.
Since reconstruction is performed only for requested rows, CP exchanges storage across the full table for computation on a small subset.

The key challenge is to construct a shared reconstruction rule that accurately represents diverse learned embeddings using short codes.
We separate each row's magnitude from its contents and define the normalized vector $z_i = \sqrt{d}\,e_i/\|e_i\|_2$.
For a code rate $b$, let $\mathcal{C}_b$ denote the set of admissible codes.
A natural encoding objective is to select a code whose reconstruction best approximates the normalized row:
\begin{equation}
    c_i^*
    =
    \arg\min_{c \in \mathcal{C}_b}
    \|z_i - f(c)\|_2^2.
    \label{eq:code-selection}
\end{equation}
At low bit rates, the quality of this approximation depends on how well the shared reconstruction rule captures the statistical properties of the source.
This motivates a source model for the normalized memory rows.

\subsection{Gaussian Source Modeling}
\label{sec:method-source}

Conditional-memory embeddings are learned through repeated updates from diverse inputs and contexts.
In hashed n-gram memories, for example, multiple n-grams may map to the same row, allowing different training signals to contribute to a shared representation \citep{cheng2026engram,liu2026scalingembeddings}.
When such contributions are distributed across many coordinates, an approximately isotropic direction provides a useful intuition for the resulting representation.
In high dimensions, isotropically distributed directions exhibit approximately Gaussian coordinate statistics after normalization \citep{diaconis1987dozen}.

Motivated by this intuition, we model the normalized row contents using a common Gaussian source while keeping row magnitudes separate.
Specifically, we approximate the coordinate distribution of $z_i$ by a standard Gaussian distribution $\mathcal{N}(0,1)$.
This source model provides a common statistical basis for representing different rows through a shared reconstruction rule, while row-specific codes capture their individual contents.

\subsection{Encoding and Norm Restoration}
\label{sec:method-encoding}

To encode the Gaussian source, we instantiate the shared reconstruction rule using trellis-coded quantization (TCQ) \citep{marcellin1990tcq,mao2010stationary}.
TCQ represents a sequence of values through a path of state transitions, allowing reconstruction values to be selected jointly rather than quantizing each coordinate independently.
We use a Gaussian HYB bitshift trellis with tail-biting.
For each row, the encoder searches for a low-error path through the shared trellis and stores its transition symbols as $c_i$.
The reconstruction function $f$ recovers the corresponding vector from these symbols using a small shared lookup table and arithmetic operations.
All rows use the same coding rule and code rate, without an additional coordinate rotation.

Although the shared source model describes normalized row contents, the magnitude of each embedding also affects the signal passed to the model.
A least-squares reconstruction scale can shrink the decoded vector when its direction differs from the original row.
CP instead restores the original row norm by setting $\alpha_i = \|e_i\|_2/\|f(c_i)\|_2$.
This preserves the magnitude of the learned representation, even when doing so increases coordinate reconstruction error.
The scale is computed once during encoding and stored with the corresponding row.

\subsection{Independent Storage and Selective Execution}
\label{sec:method-execution}

CP stores each row as an independently addressable record containing its compressed code and local scale.
For a row of dimension $d$ encoded at $b$ bits per coordinate, storing one 16-bit scale gives a rate of $b+16/d$ bits per parameter before alignment.
The shared reconstruction rule incurs only a table-wide storage cost.
Because each record can be addressed independently, compressed rows can be fetched and cached without accessing unrelated entries.

At inference time, the original address function determines the requested rows $A(x)$.
For each address $i \in A(x)$, CP fetches the corresponding code and scale, reconstructs the embedding using Equation~\ref{eq:cp-reconstruction}, and passes it to the existing readout in the original lookup order.
A fused kernel performs fetching and reconstruction for the selected records without decoding unrelated rows.
The address function, backbone parameters, and downstream computation remain unchanged.

\section{Experiments}
\label{sec:experiments}

We evaluate how well CP preserves model quality under reduced table storage.
We first compare downstream task performance at low bit rates, then examine the trade-off between storage and output-distribution fidelity across a wider range of rates.

\subsection{Experimental Setup}
\label{sec:setup}

\paragraph{Models.}
We evaluate Conditional Program (CP) on three models
that employ n-gram lookup tables: DeepSeek, Qwen,
and LongCat, as described in Section~\ref{sec:bg-memories}.
To isolate the effect of compression, we compress only
the lookup tables while keeping all other model
components unchanged.

\paragraph{Benchmarks.}
To assess the impact of compression across different
model capabilities, we evaluate 40 tasks spanning
five categories: commonsense, knowledge, reading
comprehension, multilingual understanding, and math
and code.
Beyond knowledge-intensive benchmarks, we include
GSM8K and HumanEval to evaluate mathematical reasoning
and code generation, both of which require generated
answers.
Table~\ref{tab:quality} presents seven representative
benchmark columns from this suite.

\paragraph{Baselines.}
We first compare against the original lookup tables
and table removal, where every lookup returns zero,
to assess the contribution of the stored entries.
For quantization baselines, we include round-to-nearest
(RTN), which independently rounds values to the nearest
quantization level, and NVFP4 and MXFP4, which use
four-bit floating-point formats
\citep{rouhani2023microscaling}.
To further examine the benefits of CP's design,
we adapt the trellis quantization of QTIP
\citep{tseng2024qtip}, originally developed for
dense weights, to lookup tables.
Specifically, we construct a trellis-based baseline
with within-row rotation and a global scale shared
across the table, allowing us to distinguish the
benefits of trellis quantization from those of
CP's row-wise magnitude separation.

\subsection{Results}
\label{sec:main-results}

\begin{table*}[t]
\centering\small
\caption{\textbf{Lookup-table storage and downstream performance.} Bits denotes code bits per value; table GB includes scales and offsets. Scores are on a 0--100 scale; Belebele averages eight languages. Overall equally weights the five task-group means. Evaluation details and all 40 tasks appear in Appendices~\ref{app:repro} and~\ref{app:full}.}
\label{tab:quality}\label{tab:main}\label{tab:tasks}
\setlength{\tabcolsep}{2.5pt}
\renewcommand{\arraystretch}{1.08}
\begin{tabularx}{\textwidth}{@{}Xrrrrrrrrrr@{}}
\toprule
& & & \multicolumn{8}{c}{Task score $\uparrow$} \\
\cmidrule(l){4-11}
Method & Bits & Table GB $\downarrow$ & PopQA & TriviaQA & NQ & MMLU & Belebele & GSM8K & HumanEval & \textbf{Overall} \\
\midrule
\multicolumn{11}{@{}l}{\textbf{DeepSeek-V4.1-Flash}} \\
Original & --- & 202.7 & 44.2 & 87.9 & 43.2 & 88.2 & 93.9 & 96.4 & 97.6 & 79.8 \\
w/o memory & --- & 0.0 & 20.9 & 75.0 & 30.4 & 82.2 & 89.1 & 95.8 & 90.2 & 73.5 \\
\addlinespace[3pt]
\multirow{2}{*}{RTN} & 3 & 86.0 & 43.9 & 87.9 & 43.5 & 88.3 & 93.9 & 96.8 & 98.2 & 79.9 \\
 & 2 & 61.4 & 43.1 & 87.8 & 43.1 & 88.7 & 94.1 & 97.0 & 96.3 & 79.7 \\
QTIP & 2 & 49.2 & 43.8 & 88.2 & 42.9 & 89.5 & 94.0 & 97.6 & 96.3 & 79.8 \\
\addlinespace[3pt]
\rowcolor{black!5}
 & {2} & 50.7 & 43.7 & 87.7 & 43.3 & 88.5 & 94.1 & 96.4 & 97.6 & 79.9 \\
\rowcolor{black!5}
\multirow{-2}{*}{\textbf{CP}} & 1.5 & 38.4 & 42.3 & 87.9 & 42.9 & 88.4 & 94.2 & 97.0 & 95.7 & 79.5 \\
\midrule
\multicolumn{11}{@{}l}{\textbf{Qwen3.8-Flash-Next}} \\
Original & --- & 102.4 & 36.3 & 82.3 & 42.4 & 86.8 & 93.9 & 87.6 & 95.7 & 75.6 \\
w/o memory & --- & 0.0 & 26.6 & 73.4 & 37.6 & 85.9 & 91.9 & 83.0 & 94.5 & 72.4 \\
\addlinespace[3pt]
\multirow{2}{*}{RTN} & 3 & 21.8 & 36.4 & 82.0 & 41.9 & 87.2 & 93.8 & 87.4 & 95.7 & 75.5 \\
 & 2 & 15.4 & 35.1 & 82.0 & 41.7 & 86.3 & 93.9 & 86.4 & 95.7 & 75.4 \\
QTIP & 2 & 12.8 & 36.1 & 82.4 & 42.1 & 87.3 & 93.9 & 86.6 & 95.1 & 75.4 \\
\addlinespace[3pt]
\rowcolor{black!5}
 & {2} & 13.4 & 36.4 & 82.2 & 42.1 & 86.8 & 94.0 & 86.8 & 95.7 & 75.5 \\
\rowcolor{black!5}
\multirow{-2}{*}{\textbf{CP}} & 1.5 & 10.2 & 35.4 & 82.2 & 42.6 & 86.2 & 94.0 & 86.0 & 95.7 & 75.4 \\
\midrule
\multicolumn{11}{@{}l}{\textbf{LongCat-Flash-Lite}} \\
Original & --- & 62.8 & 26.6 & 68.5 & 26.2 & 80.8 & 85.7 & 90.8 & 68.9 & 68.2 \\
w/o memory & --- & 0.0 & 11.8 & 31.2 & 10.4 & 28.5 & 25.9 & 13.0 & 9.1 & 30.5 \\
\addlinespace[3pt]
\multirow{2}{*}{RTN} & 3 & 13.7 & 26.4 & 68.5 & 26.2 & 80.7 & 85.4 & 90.6 & 66.5 & 67.9 \\
 & 2 & 9.8 & 26.1 & 67.5 & 26.3 & 79.4 & 85.1 & 88.8 & 59.1 & 66.7 \\
QTIP & 2 & 7.9 & 26.4 & 69.1 & 26.6 & 81.4 & 86.3 & 90.2 & 63.4 & 67.4 \\
\addlinespace[3pt]
\rowcolor{black!5}
 & {2} & 8.1 & 26.1 & 68.3 & 26.5 & 80.5 & 85.8 & 89.0 & 63.4 & 67.3 \\
\rowcolor{black!5}
\multirow{-2}{*}{\textbf{CP}} & 1.5 & 6.1 & 25.8 & 68.2 & 26.4 & 80.6 & 85.4 & 88.2 & 60.4 & 67.0 \\
\bottomrule
\end{tabularx}
\end{table*}

\paragraph{Benchmark Performance.}
We first examine whether CP can substantially reduce
lookup-table storage while preserving model performance.
As shown in Table~\ref{tab:quality}, CP compresses
the original tables by approximately 4 to 8 times
while retaining most of their benchmark performance.
In particular, CP largely preserves performance on
knowledge-intensive tasks where removing the lookup
table causes substantial degradation.
Compared with existing quantization methods, CP
outperforms RTN on the knowledge benchmarks for
Qwen and LongCat, while achieving comparable
performance to QTIP in the paired evaluations.
Reducing the rate further to 1.5 bits provides
additional storage savings, although some tasks
exhibit greater performance loss.

\begin{figure*}[t]
\centering
\definecolor{oiBlue}{RGB}{0,114,178}
\definecolor{oiOrange}{RGB}{230,159,0}
\definecolor{oiGreen}{RGB}{0,158,115}
\definecolor{oiPurple}{RGB}{204,121,167}
\pgfplotsset{
  ours/.style={color=oiBlue, thick, mark=*, mark size=1.8pt,
    nodes near coords, point meta=explicit symbolic, visualization depends on={value \thisrow{anchor}\as\lanchor},
    every node near coord/.style={font=\tiny, color=oiBlue, anchor=\lanchor, inner sep=1.5pt}},
  rtn/.style={color=oiOrange, densely dashed, mark=square*, mark size=1.8pt, mark options={solid}},
  qtip/.style={color=oiGreen, only marks, mark=triangle*, mark size=2.6pt},
  fp4/.style={color=oiPurple, only marks, mark=diamond*, mark size=2.6pt},
}
\begin{tikzpicture}
\begin{groupplot}[
  group style={group size=3 by 1, horizontal sep=0.9cm, y descriptions at=edge left},
  width=0.36\textwidth, height=4.0cm,
  ymode=log, ymin=0.0061, ymax=0.16,
  ytick={0.02,0.03,0.05,0.1}, yticklabels={.02,.03,.05,.10},
  log ticks with fixed point,
  xlabel={Table size (GB)},
  enlarge x limits=0.12,
  tick label style={font=\scriptsize}, label style={font=\small}, title style={font=\small},
  grid=major, grid style={gray!20},
  legend style={font=\scriptsize, draw=gray!50, /tikz/every even column/.append style={column sep=8pt}},
  legend cell align=left,
]
\nextgroupplot[title={DeepSeek-V4.1-Flash}, ylabel={Output KL}, legend to name=frontierlegend, legend columns=4]
  \addplot[ours] table[x=gb, y=gmkl, meta=bpp, col sep=comma] {figures/data/frontier_ds_ours.csv};
  \addplot[rtn] table[x=gb, y=gmkl, col sep=comma] {figures/data/frontier_ds_rtn.csv};
  \addplot[qtip] table[x=gb, y=gmkl, col sep=comma] {figures/data/frontier_ds_qtip.csv};
  \addplot[fp4] table[x=gb, y=gmkl, meta=name, col sep=comma] {figures/data/frontier_ds_fp4.csv};
  \legend{CP, RTN 2-/3-bit, QTIP 2.0 (one scale per table), NVFP4 / MXFP4}
\nextgroupplot[title={Qwen3.8-Flash-Next}]
  \addplot[ours] table[x=gb, y=gmkl, meta=bpp, col sep=comma] {figures/data/frontier_qwen_ours.csv};
  \addplot[rtn] table[x=gb, y=gmkl, col sep=comma] {figures/data/frontier_qwen_rtn.csv};
  \addplot[qtip] table[x=gb, y=gmkl, col sep=comma] {figures/data/frontier_qwen_qtip.csv};
  \addplot[fp4] table[x=gb, y=gmkl, meta=name, col sep=comma] {figures/data/frontier_qwen_fp4.csv};
\nextgroupplot[title={LongCat-Flash-Lite}]
  \addplot[ours] table[x=gb, y=gmkl, meta=bpp, col sep=comma] {figures/data/frontier_longcat_ours.csv};
  \addplot[rtn] table[x=gb, y=gmkl, col sep=comma] {figures/data/frontier_longcat_rtn.csv};
  \addplot[qtip] table[x=gb, y=gmkl, col sep=comma] {figures/data/frontier_longcat_qtip.csv};
  \addplot[fp4] table[x=gb, y=gmkl, meta=name, col sep=comma] {figures/data/frontier_longcat_fp4.csv};
\end{groupplot}
\end{tikzpicture}\\[2pt]
\pgfplotslegendfromname{frontierlegend}
\caption{Table size versus six-domain geometric-mean output KL (log scale); lower and further left is better. CP labels include per-row scales. Dashed lines connect RTN settings. Purple markers: MXFP4 (left), NVFP4 (right), including block scales. Full values and paired comparisons are in Table~\ref{tab:main_full}.}
\label{fig:frontier}
\end{figure*}
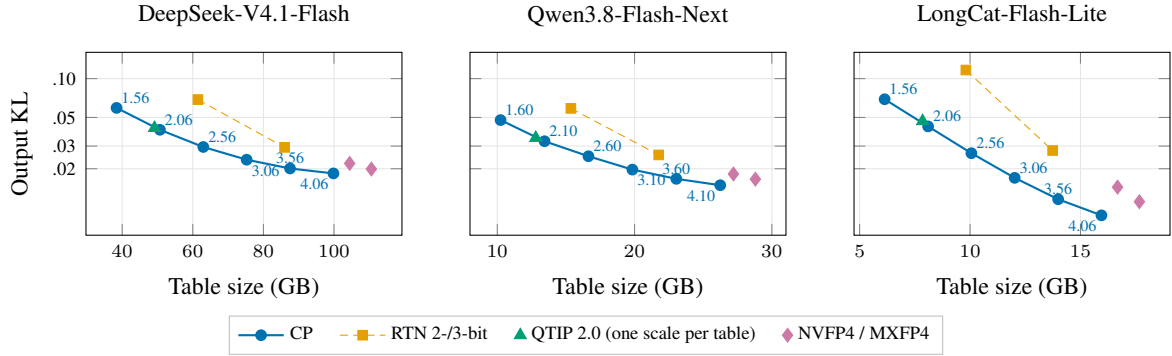

\paragraph{Functional Fidelity.}
\label{sec:frontier}
Benchmark scores alone may not fully capture how
compression changes the behavior of a model.
We therefore examine whether compressed lookup tables
preserve the original model's output distributions
by measuring per-token
$\KL(p_{\mathrm{orig}}\,\|\,p_{\mathrm{quant}})$.
We evaluate six text domains covering web, code,
mathematics, chat, and multilingual data, and
aggregate their KL values using the geometric mean
to weight each domain equally on the logarithmic scale.
As shown in Figure~\ref{fig:frontier}, at two bits,
CP reduces KL divergence by 41 to 63\% compared
with RTN and achieves a further 3 to 9\% reduction
over QTIP, with the latter improvement requiring
an additional bf16 scale per row.
The storage efficiency of CP also extends to
higher bit rates.
At 2.5 bits, CP approaches the fidelity of RTN
3-bit with less storage, while CP 3-bit further
improves fidelity.
At higher rates, CP also achieves lower KL divergence
than four-bit floating-point formats while requiring
less storage.

\subsection{Analysis}
\label{sec:analysis}
\label{sec:characterization}

We further analyze the key design choices underlying
Conditional Program. Specifically, we examine
the effect of representation design and row-wise
scaling on compression quality.

\begin{table}[t]
\centering\small
\caption{\textbf{Shared directions in frequently read rows.} Energy captured by 16 fitted directions on the 4,096 most-read held-out rows. The Gaussian reference is $16/d$.}
\label{tab:shared-directions}
\renewcommand{\arraystretch}{1.12}
\setlength{\tabcolsep}{2pt}
\begin{tabular}{@{}lrr@{}}
\toprule
2-gram table & Energy (\%) & Gaussian ref. \\
\midrule
DeepSeek, layer 1  & 19.1\% & 6.25\% \\
DeepSeek, layer 14 & 21.1\% & 6.25\% \\
Qwen               & 33.3\% & 10.0\% \\
LongCat            & 10.3\% & 6.25\% \\
\bottomrule
\end{tabular}
\end{table}

\begin{table}[t]
\centering\small
\caption{\textbf{Encoding at matched storage rates.} Output KL on Qwen, evaluated on FineWeb-Edu. VQ labels denote codebook size $\times$ block dimension.}
\label{tab:design}
\setlength{\tabcolsep}{3pt}
\renewcommand{\arraystretch}{1.12}
\begin{tabularx}{\columnwidth}{@{}Xrr@{}}
\toprule
Representation & bpp & KL $\downarrow$ \\
\midrule
\rowcolor{black!5}Shared trellis & 2.10 & 0.0198 \\
Fitted VQ, $256\times4$ & 2.10 & 0.0238 \\
Fitted VQ, $65{,}536\times8$ & 2.10 & 0.0226 \\
Mean + trellis & 2.10 & 0.0193 \\
\midrule
Mean + four PCs + trellis & 2.50 & 0.0188 \\
Trellis reference$^{\dagger}$ & 2.50 & 0.0144 \\
\bottomrule
\end{tabularx}
\par\smallskip\raggedright\footnotesize
$^{\dagger}$Interpolated on the trellis rate--KL curve. Rates include scales and coefficients; protocol in Appendix~\ref{app:eg-protocol}.
\end{table}

\paragraph{Effect of Representation Design.}
\label{sec:stats}
\label{sec:tools}
Although the normalized embedding coordinates are
approximately Gaussian, frequently retrieved rows
may still share structural information that could
benefit from a learned representation.
To investigate this possibility, we first analyze
the principal directions of frequently accessed
rows and find substantial shared structure beyond
the Gaussian reference, as shown in
Table~\ref{tab:shared-directions}.
We then compare CP's shared trellis with vector
quantization (VQ), which learns codebooks directly
from the original memory tables.
Despite the observed shared structure, CP achieves
lower row reconstruction error across all three
models and better output fidelity than VQ at
the same storage rate.
We additionally examine explicit mean and
principal-component representations, but their
improvements are limited relative to the storage
required.
These findings support using a shared trellis
rather than fitting and storing table-specific
representations.

\begin{table}[t]
\centering\small
\caption{\textbf{Rotation and row scaling.} Geometric-mean output KL over six domains with two-bit codes.}
\label{tab:scale}\label{tab:ablation}
\setlength{\tabcolsep}{2pt}
\renewcommand{\arraystretch}{1.12}
\begin{tabularx}{\columnwidth}{@{}Xlrrr@{}}
\toprule
& & \multicolumn{3}{c}{Output KL $\downarrow$} \\
\cmidrule(l){3-5}
Rotate & Scale & DeepSeek & Qwen & LongCat \\
\midrule
Yes & Global & 0.0414 & 0.0347 & 0.0468 \\
No & Global & 0.0416 & 0.0346 & 0.0473 \\
\midrule
No & Row LS & 0.0404 & 0.0349 & 0.0462 \\
\rowcolor{black!5}
No & Row norm & 0.0401 & 0.0327 & 0.0426 \\
\bottomrule
\end{tabularx}
\par\smallskip\raggedright\footnotesize
Row LS and row norm use identical codes and storage. Global: 2.00 bpp; row: 2.06/2.10/2.06 bpp. See Appendix~\ref{app:scale}.
\end{table}

\paragraph{Effect of Row-wise Scaling.}
\label{sec:propagation}
We next examine whether minimizing embedding
reconstruction error also best preserves model
behavior.
To isolate the effect of scaling, we compare
a global scale, a row-wise least-squares scale
that minimizes reconstruction error, and a
row-wise scale that restores the original norm.
The latter two use identical trellis codes and
storage budgets, differing only in their
scale values.
As shown in Table~\ref{tab:scale}, norm restoration
consistently achieves lower output KL divergence
across all three models, indicating that minimizing
row reconstruction error does not necessarily
preserve model behavior most effectively.
We also find no statistically resolved benefit
from within-row rotation, while frequency-based
bit allocation does not consistently generalize
across text domains.
These results support CP's use of norm restoration,
without rotation, and uniform bit allocation.

Additional experimental protocols, complete
benchmark results, and ablations are provided
in the Appendix.

\section{Serving Efficiency}
\label{sec:placement}\label{sec:bg-tiers}
\begin{table*}[t]
\centering\small
\caption{\textbf{Memory and workload capacity on DeepSeek.} Backbone and lookup tables reside on H200 GPUs. Four-GPU timings report mean $\pm$ standard deviation over three runs.}
\label{tab:capacity}
\renewcommand{\arraystretch}{1.14}
\setlength{\tabcolsep}{5pt}
\begin{tabularx}{\textwidth}{@{}p{.34\textwidth}*{3}{>{\raggedleft\arraybackslash}X}@{}}
\toprule
\multicolumn{4}{@{}l}{\textbf{(a) Four GPUs: memory and latency $\downarrow$}} \\
\addlinespace[2pt]
Metric / workload & FP8 & NVFP4 & \textbf{CP 2-bit} \\
\midrule
Table memory (GB/GPU) & 50.7 & 27.6 & 12.7 \\
\midrule
Prefill, $8\times4$k (s) & 64.66 $\pm$ 2.49 & 13.26 $\pm$ 0.28 & 13.27 $\pm$ 0.22 \\
Decode, batch 8 (ms/step) & 795.7 $\pm$ 11.1 & 788.9 $\pm$ 10.0 & 792.9 $\pm$ 6.3 \\
\midrule
Decode, batch 512 (s/step) & OOM & 4.03 $\pm$ 0.03 & 4.05 $\pm$ 0.06 \\
Prefill, 64k tokens (s) & OOM & 59.81 $\pm$ 4.75 & 18.23 $\pm$ 0.01 \\
\bottomrule
\end{tabularx}
\par\medskip
\begin{tabularx}{\textwidth}{@{}p{.34\textwidth}*{3}{>{\raggedleft\arraybackslash}X}@{}}
\toprule
\multicolumn{4}{@{}l}{\textbf{(b) Eight GPUs: largest successful tested workloads}} \\
\addlinespace[2pt]
Format & Table GB/GPU $\downarrow$ & Decode batch $\uparrow$ & Prefill tokens $\uparrow$ \\
\midrule
FP8 & 25.3 & 1,024 & 96k \\
\rowcolor{black!5}
\textbf{CP 2-bit} & 6.3 & $\ge$2,048 & $\ge$128k \\
\bottomrule
\end{tabularx}
\par\smallskip\raggedright\footnotesize
(a) OOM: all three runs failed. (b) One run per setting; FP8 runs out of memory at batch 2,048 and encounters severe memory pressure at 128k prefill tokens. $\ge$ denotes the largest setting tested. Workload details appear in Appendix~\ref{app:serving}.
\end{table*}

We evaluate the benefits of Conditional Programs (CPs) for
model serving. We first measure the cost of reconstructing
selectively requested rows, then examine how the reduced
memory footprint affects serving performance and workload
capacity.

\subsection{Selective Lookup}
\label{sec:access-costs}

\paragraph{Lookup layouts.}
CP stores each row as an independent record, allowing only
the requested rows to be fetched and reconstructed.
In contrast, QTIP's tiled representation distributes each
row across multiple code fragments.
To enable selective access with this layout, we implement
a tile-window decoder that retrieves only the bit windows
needed for the requested rows rather than entire tiles.
We compare this decoder against our fused row-wise decoder
under identical lookup requests, a two-bit code rate,
and per-row scale storage.
The tiled baseline uses random payloads, while CP uses
encoded model tables, so this experiment isolates access
cost rather than reconstruction fidelity.

\paragraph{Lookup latency.}
We replay lookup requests from 4,096-token sequences
constructed from held-out FineWeb-Edu text using the
DeepSeek tokenizer on a single H200 GPU.
Each token requests 48 rows across two Engram layers.
With tables stored in GPU high-bandwidth memory (HBM),
CP completes the requests in 0.190\,ms, compared with
0.221\,ms for the tile-window decoder.
When accessing pinned host DRAM over PCIe, the corresponding
times are 6.54\,ms and 10.78\,ms.
These results demonstrate that CP's row-wise representation
supports efficient selective lookup, with particularly
large benefits when accessing tables across PCIe.
We report the mean of five independent run medians,
each measured over 100 timed calls after 20 warm-up calls.
Additional comparisons and memory-access analyses
are provided in Appendix~\ref{app:format}.

\subsection{Workload Capacity}
\label{sec:capacity}

\paragraph{Experimental setup.}
To evaluate the serving benefits of reduced memory usage,
we distribute the DeepSeek backbone and tables across
four or eight H200 GPUs.
Our representative workload processes eight sequences
with 4,096-token prompts and generates 256 tokens per
sequence.
We additionally evaluate larger workloads, including
a decode batch of 512 sequences and a prefill step
of 65,536 tokens across 32 sequences.
Table~\ref{tab:capacity} summarizes the results.

\paragraph{Four-GPU serving.}
With four GPUs, the original FP8 tables occupy
50.7\,GB per GPU, compared with 12.7\,GB for CP.
For the representative workload, CP reduces prefill
latency from 64.66\,s to 13.27\,s and enables larger
workloads that otherwise run out of memory.
Compared with the four-bit NVFP4 baseline, CP also
reduces the 64k-token prefill time from 59.81\,s
to 18.23\,s, while decode times remain similar.
These results show that CP alleviates GPU memory
pressure and improves performance for
memory-intensive serving workloads.

\paragraph{Eight-GPU serving.}
With eight GPUs, the original FP8 tables fit comfortably
for smaller workloads, and step times differ by at most
2.4\% when memory pressure is limited.
However, CP frees an additional 19.0\,GB per GPU,
allowing substantially larger workloads.
On the tested grid, the largest successful decode batch
increases from 1,024 to at least 2,048 sequences,
while prefill capacity increases from 96k to at least
128k tokens.
These results demonstrate that CP's reduced memory
footprint translates into greater serving capacity,
even when smaller workloads already fit in GPU memory.
Detailed workload configurations and measurements
are provided in Appendix~\ref{app:serving}.

Together, these results highlight two complementary
benefits of CP for model serving.
Its row-wise representation enables efficient selective
lookup, while its compact storage reduces GPU memory
pressure, improving serving performance and supporting
larger workloads.
\section{Related Work}
\label{sec:related}

\paragraph{LLM weight quantization.}
GPTQ \citep{frantar2023gptq} and AWQ \citep{lin2024awq} use calibration activations to guide rounding or scaling, while QuIP, QuIP\#, QuaRot and SpinQuant use coordinate transformations to make weights easier to quantize \citep{chee2023quip,tseng2024quipsharp,ashkboos2024quarot,liu2025spinquant}.
AQLM learns additive codebooks \citep{egiazarian2024aqlm}; QTIP combines incoherence processing with trellis coding \citep{tseng2024qtip}, and HIGGS uses Gaussian-optimal grids \citep{malinovskii2025higgs}.
CP applies Gaussian trellis coding to frozen lookup rows without rotation, restoring each row's norm and storing it independently for selective access.
Because quantization can affect factual answers even when perplexity changes little \citep{jaiswal2024compressing,namburi2023cost,dutta2024accuracy}, we evaluate absolute task scores alongside output KL.

\paragraph{Embedding-table compression.}
Row-wise low-bit quantization is established for recommender embeddings \citep{guan2019posttraining,khudia2021fbgemm}; other approaches use low-precision training, frequency-based mixed precision, or structures fixed during training \citep{yang2020mixedcache,li2023alpt,li2024mpe,shi2020compositional,yin2021ttrec,desai2022robe,tsang2023cce}.
For LLM input embeddings, CARVQ combines grouped residual VQ with a corrective MLP trained for each model \citep{gou2025carvq}.
The hashed tables studied here share rows among colliding n-grams.
CP compresses these already trained tables using their row values, with a shared decoder, no model retraining and no frequency-based bit allocation.

\paragraph{Conditional-memory structure and placement.}
TN-gram replaces hashed tables with a tensor decomposition during pretraining \citep{zhou2026tngram}; TF-Engram and CXL-based systems address memory placement and access \citep{ma2026tfengram,ma2026cxl}.
TF-Engram's hash-based compression shares slots between n-grams, whereas CP numerically compresses the rows of an existing table.
SCONE stores n-gram embeddings computed by a separate model off the accelerator \citep{yu2025scone}.
Product-key memories and memory layers retrieve rows through learned key search \citep{lample2019productkeys,berges2024memorylayers}; the tables studied here use hashes of input tokens.

\section{Conclusion}
\label{sec:conclusion}

CP compresses lookup tables with row codes, local scales and a shared decoder while preserving independent access to requested rows.
Across three hashed-memory models, two-bit codes reduce table storage by roughly four- to eightfold, with most reported task scores close to their original values despite some task- and model-dependent losses.
Under GPU memory pressure, smaller tables allow larger workloads and reduce prefill time.
When memory is sufficient, complete inference steps show little speed improvement in our implementation.

\section*{Limitations}

\paragraph{Coverage.}
Our conclusions concern post-training compression of three models with hashed lookup tables; they need not extend to other memory architectures, including token-indexed tables (Appendix~\ref{app:ple}).
Most downstream comparisons use the two primary code rates, with less task coverage at higher rates than in the KL sweep.
Small task-score differences do not establish equivalence.

\paragraph{Analysis.}
The fitted-codebook and structured-residual comparisons evaluate output KL only on Qwen, limiting their generality across models.
Some equal-budget comparisons use interpolated trellis values without uncertainty estimates.
The Gaussian source approximation does not establish how training produces the observed row statistics.

\paragraph{Deployment.}
Serving measurements use a single node and eager inference; optimized engines may spend a different fraction of time on lookup.
Some configurations have only one measurement run, and tested workload limits are not exact capacity thresholds.
Access counts come from English web text, so frequency-based allocation and caching results may change with the deployment distribution.

\bibliography{references}

\clearpage
\appendix
\raggedbottom
\renewcommand{\topfraction}{0.9}
\renewcommand{\bottomfraction}{0.8}
\renewcommand{\textfraction}{0.07}
\renewcommand{\floatpagefraction}{0.7}
\renewcommand{\dbltopfraction}{0.9}
\renewcommand{\dblfloatpagefraction}{0.7}
\setcounter{topnumber}{4}
\setcounter{bottomnumber}{2}
\setcounter{totalnumber}{6}
\setcounter{dbltopnumber}{3}
\makeatletter
\setlength{\@fptop}{0pt}
\setlength{\@dblfptop}{0pt}
\setlength{\@fpsep}{20pt}
\setlength{\@dblfpsep}{20pt}
\makeatother

\section{Evaluation and Reproducibility}
\label{app:repro}

\subsection{Models and task evaluation}
Table~\ref{tab:models} lists the dimensions and original formats of the compressed tables.
\begin{table}[!hbtp]
\centering
\footnotesize
\setlength{\tabcolsep}{3.5pt}
\resizebox{\columnwidth}{!}{%
\begin{tabular}{@{}lrrrrl@{}}
\toprule
Model & Table & Rows & Row & Rows read & Original \\
 & params & & dim & per token & format \\
\midrule
DeepSeek$^{a}$ & 196.6B & 768.0M & 256 & 48 & FP8$^{b}$ \\
Qwen & 51.2B & 320.0M & 160 & 16 & BF16 \\
LongCat & 31.4B & 122.7M & 256 & 12 & BF16 \\
\bottomrule
\end{tabular}}
\caption{Lookup tables studied: the DeepSeek-V4.1-Flash Engram tables and the Qwen3.8-Flash-Next and LongCat-Flash-Lite n-gram tables. Each token reads one row per hashed head. Original size 202.7, 102.4 and 62.8\,GB. $^{a}$Two Engram layers, each 384.0M rows, 24 rows read per token. $^{b}$e4m3 with a UE8M0 scale per 32 values.}
\label{tab:models}
\end{table}

\begin{table*}[t]\centering\small
\begin{tabular*}{\textwidth}{@{\extracolsep{\fill}}llrl@{}}\toprule
Benchmark & Evaluation set & Items & Scoring \\\midrule
ARC-E / ARC-C & Test & 1000 / 1172 & Length-normalized likelihood \\
HellaSwag / PIQA & Validation & 1000 each & Length-normalized likelihood \\
WinoGrande / BoolQ & Validation & 1267 / 1000 & Likelihood accuracy \\
LAMBADA & English test & 1000 & Last-word token match \\
MMLU & Stratified test & 1000 & Five-shot choice accuracy \\
PopQA & Test, five popularity bins & 5000 & Lowercase answer contains-match \\
TriviaQA & No-context validation & 3000 & Normalized answer contains-match \\
Natural Questions & Open validation & 3610 & Normalized answer contains-match \\
EntityQuestions / SimpleQA & Test / verified evaluation & 3000 / 1000 & Normalized answer contains-match \\
SQuAD & v1.1 validation & 1000 & Token F1 \\
Belebele & Nine languages & 500/language & Three-shot choice accuracy \\
MKQA & Nine languages & 500/language & Normalized answer contains-match \\
MGSM & Six languages & 250/language & Exact numeric match \\
GSM8K & First 500 test items & 500 & Exact numeric match \\
HumanEval & Full test & 164 & Greedy pass@1 \\\bottomrule
\end{tabular*}
\caption{Evaluation suite. Counts and protocols are fixed across quantization settings. English Belebele and MKQA are separate tasks; the main table's Belebele score averages the eight non-English languages. QA uses three-shot prompts, with chat formatting for DeepSeek and Qwen and base prompts for LongCat. SQuAD uses zero-shot passage prompting. MGSM and GSM8K use native/base few-shot reasoning prompts or zero-shot chat instructions. All generation is greedy. Output limits are 16 tokens for short-answer QA, 24 for MKQA and 32 for SQuAD; GSM8K/MGSM use 512 tokens with chat prompts and 320 with base prompts, while HumanEval uses 768 and 384, respectively.}
\label{tab:eval_protocol}\end{table*}

The 40 tasks form five reporting groups: commonsense (7), knowledge (7), reading comprehension (2), multilingual (22), and math and code (2). 
The last group contains GSM8K and HumanEval. MGSM remains in the multilingual group, where its six languages are scored separately.
Items are frozen files sampled once, with seed 1234 (seed 0 for subsets taken from our evaluation library), and every model sees the same items. 
Log-likelihood tasks are scored on raw text for every model; all generation is greedy. 
Qwen and DeepSeek use their chat formats (Qwen with thinking disabled), and LongCat uses base few-shot prompts. 

\subsection{Output fidelity and statistics}
Each domain is tokenized with the model's own tokenizer and cut into 2,047-token chunks: 128 per domain for web, code, math, chat and multi5 (262,016 tokens), and 16 per language for the 15 languages of multi15 (491,280 tokens), 880 chunks per setting. 
Documents are normally truncated to 4,000 characters, and those under 200 characters are dropped; short multi15 language sets fall back to untruncated documents. 

\paragraph{Statistics.}
All intervals are 95\% percentile bootstrap intervals with seed 0. 
Task scores and benefit retained use 2,000 item-bootstrap replicates; KL uses 2,000 replicates over 2,047-token chunks, paired across settings for ratios; paired task differences use 10,000 item replicates. 

\subsection{Quantizers and encoding}
The trellis uses a 16-bit state, two values per step and tail-biting with two Viterbi passes. 
Its HYB table is built from a 512-entry k-means codebook of a 2-D standard Gaussian. 
The code scale factor is .95 at 2--3 bits, .90 at 3.5 bits and .85 at 4 bits on DeepSeek and Qwen; LongCat uses 1.0 at 2 bits and the same values above; at 1.5 bits the factors are 1.0 and 1.05. 
The rate is $b+16/d$\,bpp; the shared lookup table is not counted for DeepSeek and Qwen and adds about $6\times10^{-6}$\,bpp for LongCat. 
RTN stores a minimum and a scale per group, counted as $2\times16$ bits. 
NVFP4 has an e4m3 scale per 16 values and one fp32 scale per table (4.5\,bpp); MXFP4 has a power-of-two scale per 32 values (4.25\,bpp). 

\paragraph{Encoding cost and software.}
Encoding the 2-bit CP table takes about 3.1\,h on 4 H200 GPUs for DeepSeek (about 12 GPU-hours), 0.83\,h on 4 GPUs for Qwen, and 5.18\,h summed over LongCat's 12 tables. 
Software is PyTorch 2.14.0 and Transformers 5.17.0. 
The k-means table is not bit-reproducible across independent encodes, so the LS controls reuse the stored symbols of the norm-restoration tables. 
The Qwen RTN baseline keeps its scale and minimum in fp32 although its rate counts them as bf16. 

The table sizes in the main results are logical code and metadata sizes. SSD files may also require page padding (Appendix~\ref{app:format}).
The shared decoder state is negligible for these tables; omitting its few kilobytes does not change any displayed rate.
The RTN metadata convention above is a storage-accounting convention: Qwen's evaluated implementation actually retains those values in fp32.

\section{Complete Downstream Results}
\label{sec:appendix}\label{app:full}
\subsection{Absolute scores and removal controls}
Table~\ref{tab:tasks_full} gives the absolute score on every task.
Table~\ref{tab:quality} selects seven benchmarks from this suite and reports their scores on a 0--100 scale. The complete table below retains the original 0--1 scale.
Table~\ref{tab:popqa} splits PopQA by entity popularity, with the rarest entities in bin~1. 
It lets the reader check per bin where the knowledge loss of Section~\ref{sec:main-results} falls.
Pooling the binary item outcomes across all 40 DeepSeek tasks (so tasks with more items receive more weight), accuracy changes by $-0.07$ points with CP, $-0.34$ with RTN 2-bit and $-12.0$ with the table removed. 
On Qwen PopQA items answered correctly by the original model, a logistic regression relates errors introduced by CP to entity popularity (logistic coefficient $-.43$ [$-.71$, $-.12$]). 
The corresponding coefficient for the read count of the subject's rarest row is not statistically resolved ($-.15$ [$-.49$, $+.17$]). 
Every setting has scores on all 40 tasks; the table averages non-English languages within each multilingual benchmark. 
The paired comparisons of Section~\ref{sec:main-results} resample the same items for every setting of a task (10,000 replicates; Appendix~\ref{app:repro}). 
On DeepSeek, the 1.5-bit code has a lower knowledge-group score than QTIP 2.0 (difference $-0.80$ percentage points, 95\% interval [$-1.23$, $-0.38$]); the PopQA difference is $-1.50$ [$-2.14$, $-0.84$] points. 

\begin{table*}[t]
\centering
\footnotesize
\setlength{\tabcolsep}{1.4pt}
\resizebox{\textwidth}{!}{%
\begin{tabular}{@{}l rrrrrrr@{\hspace{6pt}}rrrrrrr@{\hspace{6pt}}rrrrrrr@{}}
\toprule
 & \multicolumn{7}{c}{DeepSeek-V4.1-Flash} & \multicolumn{7}{c}{Qwen3.8-Flash-Next} & \multicolumn{7}{c}{LongCat-Flash-Lite} \\
\cmidrule(lr){2-8}\cmidrule(lr){9-15}\cmidrule(lr){16-22}
& & \multicolumn{2}{c}{RTN} & & \multicolumn{2}{c}{CP} & & & \multicolumn{2}{c}{RTN} & & \multicolumn{2}{c}{CP} & & & \multicolumn{2}{c}{RTN} & & \multicolumn{2}{c}{CP} \\
\cmidrule(lr){4-5}\cmidrule(lr){7-8}\cmidrule(lr){11-12}\cmidrule(lr){14-15}\cmidrule(lr){18-19}\cmidrule(lr){21-22}
Task & \multicolumn{1}{c}{\scriptsize Orig.} & \multicolumn{1}{c}{\scriptsize Rem.} & \multicolumn{1}{c}{\scriptsize 3} & \multicolumn{1}{c}{\scriptsize 2} & \multicolumn{1}{c}{\scriptsize QTIP} & \multicolumn{1}{c}{\scriptsize 2} & \multicolumn{1}{c@{\hspace{6pt}}}{\scriptsize 1.5} & \multicolumn{1}{c}{\scriptsize Orig.} & \multicolumn{1}{c}{\scriptsize Rem.} & \multicolumn{1}{c}{\scriptsize 3} & \multicolumn{1}{c}{\scriptsize 2} & \multicolumn{1}{c}{\scriptsize QTIP} & \multicolumn{1}{c}{\scriptsize 2} & \multicolumn{1}{c@{\hspace{6pt}}}{\scriptsize 1.5} & \multicolumn{1}{c}{\scriptsize Orig.} & \multicolumn{1}{c}{\scriptsize Rem.} & \multicolumn{1}{c}{\scriptsize 3} & \multicolumn{1}{c}{\scriptsize 2} & \multicolumn{1}{c}{\scriptsize QTIP} & \multicolumn{1}{c}{\scriptsize 2} & \multicolumn{1}{c@{}}{\scriptsize 1.5} \\
\midrule
\multicolumn{22}{@{}l}{\textit{Commonsense}} \\
\bottomrule
\end{tabular}}
\caption{All task results on the 0--1 scale. Scores use the task-specific metrics in Appendix~\ref{app:repro}, including token F1 for SQuAD. Orig.: original table; Rem.: table removed. RTN codes use 3 or 2 bits; QTIP uses 2 bits; CP uses 2 or 1.5 bits. Belebele-$n$, MKQA-$n$ and MGSM-$n$ average $n$ non-English languages. Task differences and their intervals are reported in Table~\ref{tab:paired_tasks}.}
\label{tab:tasks_full}
\end{table*}

\begin{table*}[p]\centering\footnotesize
\setlength{\tabcolsep}{2pt}
\begin{tabular*}{\textwidth}{@{\extracolsep{\fill}}lrrrrrrrrrrrrrrr@{}}\toprule
 & \multicolumn{5}{c}{DeepSeek} & \multicolumn{5}{c}{Qwen} & \multicolumn{5}{c}{LongCat} \\
\cmidrule(lr){2-6}\cmidrule(lr){7-11}\cmidrule(l){12-16}
Setting & 1 & 2 & 3 & 4 & 5 & 1 & 2 & 3 & 4 & 5 & 1 & 2 & 3 & 4 & 5 \\\midrule
Original & .335 & .333 & .379 & .482 & .683 & .261 & .230 & .269 & .380 & .674 & .224 & .157 & .165 & .243 & .542 \\
Table removed & .133 & .096 & .108 & .190 & .520 & .188 & .132 & .149 & .253 & .606 & .149 & .080 & .067 & .065 & .228 \\
RTN 3-bit & .323 & .330 & .375 & .481 & .686 & .259 & .239 & .272 & .378 & .673 & .229 & .155 & .159 & .243 & .534 \\
RTN 2-bit & .310 & .312 & .367 & .470 & .694 & .245 & .215 & .254 & .376 & .663 & .228 & .153 & .153 & .239 & .534 \\
QTIP 2.0 & .323 & .332 & .370 & .478 & .689 & .254 & .232 & .262 & .380 & .676 & .224 & .159 & .155 & .246 & .537 \\
\textbf{CP, 2-bit code} & .318 & .325 & .381 & .477 & .685 & .267 & .226 & .267 & .385 & .676 & .217 & .159 & .155 & .237 & .539 \\
CP, 1.5-bit code & .300 & .307 & .358 & .466 & .686 & .253 & .216 & .260 & .372 & .669 & .213 & .151 & .152 & .240 & .533 \\
\bottomrule\end{tabular*}
\caption{PopQA accuracy by entity popularity bin (1 = rarest). Settings match Table~\ref{tab:tasks_full}.}
\label{tab:popqa}\end{table*}

\subsection{Paired comparisons}
Tables~\ref{tab:paired_ds}--\ref{tab:paired_longcat} give per-task score differences in percentage points.
All methods in a pair use the same sampled items in each bootstrap replicate.
Intervals describe individual comparisons and are not adjusted for testing many tasks.
A confidence interval containing zero does not establish that two methods are equivalent.

For the seven-task knowledge group, the two-bit representation improves on RTN by 0.67 [0.27, 1.06] points on Qwen and 0.82 [0.23, 1.41] on LongCat; DeepSeek gives 0.40 [$-0.04$, 0.84].
The corresponding comparisons with QTIP include zero, as do the PopQA and combined GSM8K/HumanEval comparisons in every model.

\begin{table*}[p]\centering\footnotesize
\setlength{\tabcolsep}{4pt}
\begin{tabular*}{\textwidth}{@{\extracolsep{\fill}}lrrrrr@{}}\toprule
Task & $n$ & CP 2 $-$ RTN 2 & CP 2 $-$ QTIP 2 & CP 1.5 $-$ QTIP 2 & CP 1.5 $-$ RTN 2 \\\midrule
ARC-E & 1000 & +0.6 [-0.7, +1.9] & +1.1 [+0.0, +2.2] & +0.2 [-1.1, +1.5] & -0.3 [-1.6, +1.0] \\
ARC-C & 1172 & -0.6 [-2.2, +1.0] & +0.0 [-1.5, +1.5] & -0.1 [-1.7, +1.5] & -0.7 [-2.5, +1.2] \\
HellaSwag & 1000 & +0.0 [-0.9, +0.9] & -0.6 [-1.4, +0.2] & +0.2 [-0.6, +1.0] & +0.8 [-0.1, +1.7] \\
PIQA & 1000 & +0.9 [-0.1, +1.9] & +0.9 [-0.1, +2.0] & +0.4 [-0.7, +1.5] & +0.4 [-0.6, +1.4] \\
WinoGrande & 1267 & +0.5 [-1.1, +2.1] & +0.3 [-1.1, +1.7] & -0.6 [-2.1, +1.0] & -0.4 [-2.1, +1.3] \\
BoolQ & 1000 & +0.1 [-0.9, +1.1] & -0.3 [-1.2, +0.6] & +0.1 [-0.9, +1.1] & +0.5 [-0.5, +1.6] \\
LAMBADA & 1000 & +0.0 [-1.3, +1.3] & +0.1 [-1.3, +1.5] & +0.1 [-1.3, +1.5] & +0.0 [-1.3, +1.3] \\
MMLU & 1000 & -0.2 [-1.1, +0.7] & -1.0 [-1.8, -0.2] & -1.1 [-2.1, -0.2] & -0.3 [-1.2, +0.6] \\
PopQA & 5000 & +0.7 [+0.0, +1.3] & -0.1 [-0.7, +0.5] & -1.5 [-2.1, -0.8] & -0.7 [-1.4, -0.1] \\
TriviaQA & 3000 & -0.1 [-0.6, +0.5] & -0.5 [-1.0, +0.0] & -0.3 [-0.8, +0.2] & +0.1 [-0.4, +0.6] \\
NQ & 3610 & +0.1 [-0.6, +0.9] & +0.3 [-0.3, +1.0] & -0.1 [-0.8, +0.7] & -0.2 [-1.0, +0.5] \\
EntityQs & 3000 & +1.0 [+0.1, +1.9] & +0.3 [-0.7, +1.2] & -1.7 [-2.6, -0.7] & -0.9 [-2.0, +0.1] \\
SimpleQA & 1000 & +2.1 [+0.5, +3.8] & +1.5 [+0.0, +3.0] & +0.4 [-0.9, +1.7] & +1.0 [-0.7, +2.7] \\
MKQA (en) & 500 & -0.8 [-2.8, +1.2] & -0.2 [-2.0, +1.4] & -1.4 [-3.4, +0.6] & -2.0 [-4.2, +0.0] \\
SQuAD & 1000 & +0.2 [-0.6, +0.9] & +1.4 [+0.8, +2.1] & +0.4 [-0.2, +1.0] & -0.9 [-1.6, -0.1] \\
Belebele (en) & 500 & +0.4 [-0.6, +1.6] & -0.2 [-0.8, +0.4] & +0.0 [-0.6, +0.6] & +0.6 [-0.2, +1.6] \\
belebele\_zho\_Hans & 500 & -0.8 [-1.8, +0.0] & -0.8 [-1.8, +0.0] & -0.8 [-1.6, -0.2] & -0.8 [-1.6, -0.2] \\
belebele\_jpn\_Jpan & 500 & -0.8 [-2.0, +0.2] & -0.8 [-1.6, -0.2] & -0.2 [-1.2, +0.8] & -0.2 [-1.2, +0.6] \\
belebele\_arb\_Arab & 500 & +0.0 [-1.2, +1.2] & -0.4 [-1.6, +0.6] & +0.6 [-0.6, +1.8] & +1.0 [-0.2, +2.4] \\
belebele\_hin\_Deva & 500 & +0.2 [-1.0, +1.6] & +0.8 [-0.4, +2.0] & +0.4 [-1.0, +1.8] & -0.2 [-1.6, +1.0] \\
belebele\_rus\_Cyrl & 500 & +0.2 [-0.6, +1.0] & +0.4 [-0.6, +1.4] & -0.4 [-1.6, +0.6] & -0.6 [-1.6, +0.4] \\
belebele\_swh\_Latn & 500 & -0.2 [-1.6, +1.2] & +0.2 [-0.8, +1.2] & +0.6 [-0.4, +1.6] & +0.2 [-0.8, +1.2] \\
belebele\_kor\_Hang & 500 & +0.8 [+0.2, +1.6] & +1.2 [+0.4, +2.2] & +0.8 [+0.2, +1.6] & +0.4 [-0.4, +1.2] \\
belebele\_deu\_Latn & 500 & +0.4 [-0.6, +1.4] & +0.2 [-0.8, +1.2] & +0.4 [-0.6, +1.4] & +0.6 [-0.4, +1.6] \\
mkqa\_zh\_cn & 500 & -0.6 [-2.4, +1.2] & -0.6 [-2.2, +1.0] & -1.6 [-3.4, +0.2] & -1.6 [-3.8, +0.4] \\
mkqa\_ja & 500 & +0.6 [-1.4, +2.6] & -0.4 [-2.2, +1.4] & -1.0 [-3.0, +1.0] & +0.0 [-2.2, +2.2] \\
mkqa\_ar & 500 & -1.0 [-3.2, +1.2] & -0.2 [-2.0, +1.6] & -0.6 [-2.6, +1.4] & -1.4 [-3.6, +0.6] \\
mkqa\_ko & 500 & -1.6 [-3.8, +0.6] & -1.0 [-3.2, +1.2] & -0.8 [-3.0, +1.2] & -1.4 [-3.6, +0.8] \\
mkqa\_ru & 500 & +0.6 [-1.6, +3.0] & +0.8 [-1.2, +2.8] & +1.2 [-0.8, +3.2] & +1.0 [-1.0, +3.2] \\
mkqa\_th & 500 & +1.6 [-0.6, +3.8] & +1.6 [-0.2, +3.4] & -0.8 [-2.6, +1.0] & -0.8 [-2.8, +1.2] \\
mkqa\_he & 500 & +0.8 [-1.0, +2.8] & +0.2 [-1.4, +1.8] & +1.0 [-1.0, +3.0] & +1.6 [-0.4, +3.6] \\
mkqa\_de & 500 & -0.2 [-2.2, +1.8] & +0.4 [-1.2, +2.0] & -0.2 [-2.0, +1.8] & -0.8 [-2.8, +1.2] \\
mgsm\_zh & 250 & -1.2 [-2.8, +0.0] & +0.0 [-2.4, +2.4] & +0.0 [-2.0, +2.0] & -1.2 [-3.2, +0.4] \\
mgsm\_ja & 250 & +0.8 [-1.2, +3.2] & +0.0 [-2.8, +2.8] & -0.4 [-2.8, +2.0] & +0.4 [-2.4, +3.2] \\
mgsm\_ru & 250 & +0.8 [-1.2, +2.8] & -0.4 [-2.8, +2.0] & -1.6 [-3.6, +0.0] & -0.4 [-2.8, +2.0] \\
mgsm\_de & 250 & -0.4 [-2.0, +0.8] & +0.4 [-1.2, +2.4] & +0.4 [-1.2, +2.4] & -0.4 [-2.0, +0.8] \\
mgsm\_sw & 250 & +0.0 [-2.4, +2.4] & -0.4 [-3.2, +2.4] & +0.0 [-2.4, +2.4] & +0.4 [-2.0, +2.8] \\
mgsm\_bn & 250 & +0.0 [-3.2, +3.2] & -2.4 [-5.6, +0.8] & -1.6 [-4.4, +1.2] & +0.8 [-2.0, +3.6] \\
GSM8K & 500 & -0.6 [-1.6, +0.4] & -1.2 [-2.4, +0.0] & -0.6 [-1.6, +0.4] & +0.0 [-1.2, +1.2] \\
HumanEval & 164 & +1.2 [+0.0, +3.0] & +1.2 [+0.0, +3.0] & -0.6 [-3.7, +1.8] & -0.6 [-3.7, +1.8] \\
\bottomrule\end{tabular*}
\caption{Paired task-score differences for DeepSeek-V4.1-Flash in percentage points, with 95\% intervals from 10,000 paired item-bootstrap replicates. Positive values favor the first method. Differences are computed before rounding, so they may differ from subtracting the displayed scores. Intervals are per comparison, without multiplicity adjustment; isolated differences across 40 tasks require caution.}
\label{tab:paired_ds}\label{tab:paired_tasks}\end{table*}

\begin{table*}[p]\centering\footnotesize
\setlength{\tabcolsep}{4pt}
\begin{tabular*}{\textwidth}{@{\extracolsep{\fill}}lrrrrr@{}}\toprule
Task & $n$ & CP 2 $-$ RTN 2 & CP 2 $-$ QTIP 2 & CP 1.5 $-$ QTIP 2 & CP 1.5 $-$ RTN 2 \\\midrule
ARC-E & 1000 & -0.9 [-2.5, +0.6] & -1.1 [-2.5, +0.3] & +0.2 [-1.3, +1.7] & +0.4 [-1.1, +1.9] \\
ARC-C & 1172 & -0.2 [-1.9, +1.5] & +0.0 [-1.7, +1.7] & -0.6 [-2.3, +1.0] & -0.8 [-2.5, +0.9] \\
HellaSwag & 1000 & +0.2 [-0.6, +1.0] & -0.1 [-0.8, +0.6] & -0.3 [-1.3, +0.6] & +0.0 [-1.0, +1.0] \\
PIQA & 1000 & -0.6 [-2.0, +0.8] & +0.8 [-0.4, +2.0] & +0.6 [-0.7, +2.0] & -0.8 [-2.2, +0.6] \\
WinoGrande & 1267 & -2.1 [-4.3, +0.2] & -0.9 [-2.8, +0.9] & +1.3 [-0.6, +3.2] & +0.2 [-1.8, +2.3] \\
BoolQ & 1000 & +0.2 [-1.4, +1.9] & +0.5 [-0.9, +2.0] & -0.3 [-1.8, +1.2] & -0.6 [-2.2, +1.0] \\
LAMBADA & 1000 & +0.6 [-0.5, +1.7] & +0.4 [-0.7, +1.5] & -0.1 [-1.3, +1.1] & +0.1 [-1.0, +1.3] \\
MMLU & 1000 & +0.5 [-0.8, +1.8] & -0.5 [-1.6, +0.6] & -1.1 [-2.3, +0.1] & -0.1 [-1.3, +1.1] \\
PopQA & 5000 & +1.4 [+0.7, +2.0] & +0.3 [-0.2, +0.9] & -0.7 [-1.2, -0.1] & +0.3 [-0.3, +1.0] \\
TriviaQA & 3000 & +0.2 [-0.4, +0.8] & -0.2 [-0.8, +0.4] & -0.2 [-0.8, +0.5] & +0.2 [-0.4, +0.9] \\
NQ & 3610 & +0.4 [-0.2, +1.1] & +0.0 [-0.6, +0.6] & +0.5 [-0.1, +1.1] & +0.9 [+0.3, +1.6] \\
EntityQs & 3000 & +1.2 [+0.2, +2.2] & -0.6 [-1.4, +0.3] & -0.7 [-1.6, +0.2] & +1.1 [+0.1, +2.0] \\
SimpleQA & 1000 & +1.2 [-0.3, +2.6] & +0.7 [-0.4, +1.8] & +0.8 [-0.3, +1.9] & +1.3 [-0.1, +2.6] \\
MKQA (en) & 500 & -0.2 [-1.6, +1.2] & +0.4 [-1.0, +1.8] & +0.2 [-1.0, +1.6] & -0.4 [-2.0, +1.2] \\
SQuAD & 1000 & +0.4 [+0.0, +0.7] & +0.2 [-0.1, +0.6] & +0.2 [-0.2, +0.6] & +0.3 [-0.1, +0.8] \\
Belebele (en) & 500 & -0.4 [-1.2, +0.4] & -0.6 [-1.4, +0.0] & +0.6 [+0.0, +1.4] & +0.8 [+0.0, +1.8] \\
belebele\_zho\_Hans & 500 & +0.0 [-1.0, +1.0] & -0.4 [-1.6, +0.6] & +0.2 [-0.4, +1.0] & +0.6 [+0.0, +1.4] \\
belebele\_jpn\_Jpan & 500 & +0.6 [-0.2, +1.6] & +0.0 [-0.8, +0.8] & -0.2 [-0.8, +0.4] & +0.4 [-0.4, +1.2] \\
belebele\_arb\_Arab & 500 & -0.4 [-1.2, +0.4] & +0.0 [-1.0, +1.0] & -0.4 [-1.4, +0.6] & -0.8 [-1.8, +0.0] \\
belebele\_hin\_Deva & 500 & +1.2 [-0.6, +3.2] & +1.6 [-0.2, +3.4] & +1.2 [-0.4, +2.8] & +0.8 [-0.8, +2.4] \\
belebele\_rus\_Cyrl & 500 & -0.4 [-1.0, +0.0] & -0.2 [-1.0, +0.4] & +0.0 [-0.8, +0.8] & -0.2 [-1.0, +0.4] \\
belebele\_swh\_Latn & 500 & -0.4 [-1.8, +1.0] & -0.6 [-1.6, +0.4] & +0.2 [-1.0, +1.4] & +0.4 [-0.6, +1.6] \\
belebele\_kor\_Hang & 500 & +0.0 [-0.8, +0.8] & -0.2 [-1.2, +0.6] & -0.2 [-1.0, +0.6] & +0.0 [-0.8, +0.8] \\
belebele\_deu\_Latn & 500 & +0.0 [-1.0, +1.0] & +0.0 [-0.8, +0.8] & +0.0 [-0.8, +0.8] & +0.0 [-1.0, +1.0] \\
mkqa\_zh\_cn & 500 & -1.8 [-3.4, -0.4] & -0.2 [-1.4, +1.0] & -1.0 [-2.4, +0.2] & -2.6 [-4.4, -1.0] \\
mkqa\_ja & 500 & -0.4 [-2.0, +1.2] & -0.8 [-2.2, +0.6] & -0.6 [-2.2, +1.0] & -0.2 [-1.8, +1.4] \\
mkqa\_ar & 500 & +0.8 [-0.6, +2.2] & +0.4 [-0.4, +1.2] & +0.0 [-1.4, +1.4] & +0.4 [-1.2, +2.0] \\
mkqa\_ko & 500 & +0.6 [-1.0, +2.2] & +0.2 [-1.4, +1.6] & -0.4 [-2.2, +1.4] & +0.0 [-1.6, +1.6] \\
mkqa\_ru & 500 & -0.8 [-2.4, +0.8] & +0.8 [-0.6, +2.2] & +0.4 [-1.2, +2.2] & -1.2 [-3.2, +0.8] \\
mkqa\_th & 500 & +0.4 [-1.2, +2.0] & +0.2 [-1.4, +1.8] & -0.4 [-1.8, +1.0] & -0.2 [-1.6, +1.2] \\
mkqa\_he & 500 & +0.0 [-1.6, +1.6] & +0.6 [-1.0, +2.2] & -0.2 [-1.6, +1.2] & -0.8 [-2.4, +0.8] \\
mkqa\_de & 500 & +0.2 [-1.4, +1.8] & +0.0 [-1.2, +1.2] & -0.2 [-1.6, +1.4] & +0.0 [-2.0, +2.0] \\
mgsm\_zh & 250 & -0.8 [-4.4, +2.8] & -0.4 [-4.4, +3.6] & +1.2 [-2.4, +4.8] & +0.8 [-2.8, +4.4] \\
mgsm\_ja & 250 & +2.4 [-1.6, +6.8] & +4.0 [-0.4, +8.4] & -0.4 [-4.4, +3.6] & -2.0 [-5.2, +1.2] \\
mgsm\_ru & 250 & +0.0 [-3.2, +3.2] & +3.2 [-0.4, +6.8] & +0.8 [-2.4, +4.0] & -2.4 [-6.0, +1.2] \\
mgsm\_de & 250 & +1.2 [-1.6, +4.0] & -2.4 [-5.6, +0.8] & -2.0 [-5.6, +1.6] & +1.6 [-2.4, +5.6] \\
mgsm\_sw & 250 & +2.0 [-2.4, +6.4] & +1.2 [-2.8, +5.2] & -1.6 [-5.6, +2.4] & -0.8 [-5.6, +4.0] \\
mgsm\_bn & 250 & +2.0 [-2.4, +6.4] & +2.8 [-1.6, +7.2] & +0.0 [-4.8, +4.8] & -0.8 [-5.2, +3.6] \\
GSM8K & 500 & +0.4 [-1.2, +2.0] & +0.2 [-1.6, +2.0] & -0.6 [-2.6, +1.4] & -0.4 [-2.4, +1.6] \\
HumanEval & 164 & +0.0 [+0.0, +0.0] & +0.6 [-1.2, +3.0] & +0.6 [-1.2, +3.0] & +0.0 [+0.0, +0.0] \\
\bottomrule\end{tabular*}
\caption{Paired task-score differences for Qwen3.8-Flash-Next in percentage points, with 95\% intervals from 10,000 paired item-bootstrap replicates. Positive values favor the first method. Differences are computed before rounding, so they may differ from subtracting the displayed scores. Intervals are per comparison, without multiplicity adjustment; isolated differences across 40 tasks require caution.}
\label{tab:paired_qwen}\end{table*}

\begin{table*}[p]\centering\footnotesize
\setlength{\tabcolsep}{4pt}
\begin{tabular*}{\textwidth}{@{\extracolsep{\fill}}lrrrrr@{}}\toprule
Task & $n$ & CP 2 $-$ RTN 2 & CP 2 $-$ QTIP 2 & CP 1.5 $-$ QTIP 2 & CP 1.5 $-$ RTN 2 \\\midrule
ARC-E & 1000 & +0.5 [-1.0, +2.0] & +0.1 [-1.2, +1.4] & +0.2 [-1.2, +1.6] & +0.6 [-0.9, +2.1] \\
ARC-C & 1172 & +2.1 [+0.4, +3.8] & +1.3 [-0.2, +2.7] & +1.0 [-0.5, +2.6] & +1.9 [+0.2, +3.6] \\
HellaSwag & 1000 & +1.0 [-0.3, +2.4] & +1.1 [-0.1, +2.3] & -0.6 [-2.0, +0.7] & -0.7 [-2.1, +0.6] \\
PIQA & 1000 & +0.9 [-0.8, +2.6] & -1.2 [-2.7, +0.3] & -1.4 [-2.9, +0.1] & +0.7 [-0.8, +2.3] \\
WinoGrande & 1267 & +0.4 [-2.3, +3.1] & +0.4 [-2.0, +2.7] & +1.1 [-1.4, +3.6] & +1.1 [-1.5, +3.8] \\
BoolQ & 1000 & +1.4 [+0.0, +2.9] & +0.7 [-0.5, +1.9] & +0.7 [-0.6, +2.0] & +1.4 [-0.1, +2.9] \\
LAMBADA & 1000 & +2.8 [+1.1, +4.5] & +0.7 [-0.3, +1.8] & +0.3 [-1.0, +1.6] & +2.4 [+0.8, +4.0] \\
MMLU & 1000 & +1.1 [-0.4, +2.6] & -0.9 [-2.2, +0.4] & -0.8 [-2.2, +0.5] & +1.2 [-0.2, +2.6] \\
PopQA & 5000 & +0.0 [-0.7, +0.7] & -0.3 [-0.9, +0.3] & -0.6 [-1.3, +0.0] & -0.4 [-1.0, +0.3] \\
TriviaQA & 3000 & +0.8 [-0.2, +1.7] & -0.8 [-1.6, -0.0] & -0.9 [-1.7, -0.1] & +0.7 [-0.2, +1.6] \\
NQ & 3610 & +0.1 [-0.7, +1.0] & -0.1 [-0.9, +0.7] & -0.2 [-1.0, +0.7] & +0.1 [-0.9, +1.0] \\
EntityQs & 3000 & +0.6 [-0.4, +1.6] & -0.4 [-1.3, +0.5] & +0.1 [-0.9, +1.1] & +1.1 [+0.0, +2.1] \\
SimpleQA & 1000 & +0.3 [-1.2, +1.8] & -0.5 [-1.9, +0.8] & -0.8 [-2.2, +0.6] & +0.0 [-1.5, +1.5] \\
MKQA (en) & 500 & +2.8 [-0.2, +5.8] & +2.6 [+0.2, +5.0] & +0.4 [-2.2, +3.0] & +0.6 [-2.4, +3.6] \\
SQuAD & 1000 & -3.7 [-4.8, -2.5] & -0.1 [-1.0, +0.8] & +2.2 [+1.2, +3.3] & -1.4 [-2.4, -0.4] \\
Belebele (en) & 500 & +0.2 [+0.0, +0.6] & +0.2 [-0.6, +1.0] & +0.0 [-0.6, +0.6] & +0.0 [-0.8, +0.8] \\
belebele\_zho\_Hans & 500 & +0.6 [-0.8, +2.0] & +0.4 [-0.6, +1.6] & -0.2 [-1.4, +1.0] & +0.0 [-1.6, +1.6] \\
belebele\_jpn\_Jpan & 500 & -0.8 [-2.4, +0.8] & -1.2 [-2.8, +0.2] & -0.8 [-2.2, +0.6] & -0.4 [-2.2, +1.4] \\
belebele\_arb\_Arab & 500 & +0.6 [-0.6, +1.8] & -1.2 [-2.2, -0.4] & -1.0 [-2.2, +0.0] & +0.8 [-0.6, +2.2] \\
belebele\_hin\_Deva & 500 & +2.0 [-0.6, +4.6] & +0.0 [-2.2, +2.2] & -0.6 [-2.8, +1.8] & +1.4 [-1.2, +4.0] \\
belebele\_rus\_Cyrl & 500 & -0.4 [-1.6, +0.8] & -0.6 [-1.6, +0.4] & -1.4 [-2.8, -0.2] & -1.2 [-2.6, +0.2] \\
belebele\_swh\_Latn & 500 & +3.2 [+0.2, +6.2] & -0.8 [-3.6, +2.0] & -2.2 [-5.2, +0.8] & +1.8 [-1.6, +5.2] \\
belebele\_kor\_Hang & 500 & -0.6 [-2.2, +1.0] & -0.2 [-1.4, +1.0] & -0.8 [-2.4, +0.8] & -1.2 [-3.0, +0.6] \\
belebele\_deu\_Latn & 500 & +1.0 [-0.6, +2.8] & -0.8 [-2.0, +0.4] & -0.6 [-1.8, +0.4] & +1.2 [-0.4, +3.0] \\
mkqa\_zh\_cn & 500 & +2.0 [+0.0, +4.0] & +0.2 [-1.4, +1.8] & -0.2 [-2.0, +1.6] & +1.6 [-0.4, +3.6] \\
mkqa\_ja & 500 & -0.8 [-2.0, +0.4] & -0.6 [-1.8, +0.6] & -0.2 [-1.4, +1.0] & -0.4 [-1.8, +1.0] \\
mkqa\_ar & 500 & -0.4 [-1.8, +1.0] & -0.4 [-1.4, +0.6] & +0.4 [-0.6, +1.6] & +0.4 [-1.0, +1.8] \\
mkqa\_ko & 500 & +0.8 [-0.6, +2.2] & +0.8 [-0.4, +2.2] & +0.6 [-1.0, +2.2] & +0.6 [-0.6, +1.8] \\
mkqa\_ru & 500 & -0.2 [-2.2, +1.8] & -0.4 [-2.4, +1.6] & -0.2 [-2.0, +1.6] & +0.0 [-2.4, +2.4] \\
mkqa\_th & 500 & -0.4 [-1.6, +0.6] & -0.2 [-1.4, +1.0] & +0.6 [-0.6, +1.8] & +0.4 [-0.6, +1.4] \\
mkqa\_he & 500 & +0.4 [-0.6, +1.6] & +0.4 [-0.8, +1.6] & +0.4 [-0.4, +1.2] & +0.4 [-0.8, +1.6] \\
mkqa\_de & 500 & +0.6 [-1.8, +2.8] & -0.8 [-3.0, +1.4] & -2.2 [-4.4, +0.0] & -0.8 [-3.0, +1.4] \\
mgsm\_zh & 250 & -2.0 [-6.0, +2.0] & -2.8 [-6.0, +0.4] & -0.4 [-3.6, +2.4] & +0.4 [-2.8, +3.6] \\
mgsm\_ja & 250 & +0.8 [-3.6, +4.8] & +2.4 [-2.0, +6.8] & +1.6 [-2.8, +6.0] & +0.0 [-4.0, +4.0] \\
mgsm\_ru & 250 & -0.8 [-4.4, +2.8] & +2.0 [-1.2, +5.6] & +2.8 [-0.4, +6.4] & +0.0 [-3.6, +3.6] \\
mgsm\_de & 250 & +0.4 [-3.6, +4.4] & -2.4 [-6.0, +1.2] & -2.0 [-5.6, +1.2] & +0.8 [-2.4, +4.0] \\
mgsm\_sw & 250 & +2.4 [-4.0, +8.8] & -1.2 [-6.8, +4.4] & +0.0 [-6.0, +6.0] & +3.6 [-2.8, +9.6] \\
mgsm\_bn & 250 & +0.0 [-5.2, +5.2] & +0.8 [-4.0, +5.6] & -2.0 [-7.2, +3.2] & -2.8 [-8.0, +2.4] \\
GSM8K & 500 & +0.2 [-2.0, +2.4] & -1.2 [-3.4, +1.0] & -2.0 [-4.2, +0.0] & -0.6 [-2.8, +1.6] \\
HumanEval & 164 & +4.3 [-2.4, +11.0] & +0.0 [-5.5, +5.5] & -3.0 [-9.1, +3.0] & +1.2 [-6.1, +8.5] \\
\bottomrule\end{tabular*}
\caption{Paired task-score differences for LongCat-Flash-Lite in percentage points, with 95\% intervals from 10,000 paired item-bootstrap replicates. Positive values favor the first method. Differences are computed before rounding, so they may differ from subtracting the displayed scores. Intervals are per comparison, without multiplicity adjustment; isolated differences across 40 tasks require caution.}
\label{tab:paired_longcat}\end{table*}

\subsection{Benefit retained as a secondary measure}
For a task score $s$, the fraction of the table's measured benefit retained is
\[
R=\frac{s_q-s_{\mathrm{rem}}}{s_{\mathrm{orig}}-s_{\mathrm{rem}}}.
\]
Here $s_q$, $s_{\mathrm{rem}}$ and $s_{\mathrm{orig}}$ use the compressed, removed and original table.
For a group, we sum the score gains in the numerator and denominator over its tasks before taking the ratio.
This measures preservation of the removal gap; it is not accuracy relative to the original model.
It can exceed one and becomes unstable when the original and removed scores are close.
We use absolute scores and paired differences in the main comparison for that reason.
We report perplexity in its original units in Appendix~\ref{app:fidelity}.

\clearpage
\makeatletter
\newenvironment{appendixfixedtable}
  {\par\addvspace{8pt}\noindent\begin{minipage}{\linewidth}\def\@captype{table}}
  {\end{minipage}\par\addvspace{8pt}}
\makeatother
\twocolumn[{%
\begin{minipage}{\textwidth}
\section{Full Rate--Distortion Results}
\label{app:fidelity}
\begingroup
\renewenvironment{table*}[1][]{\begin{appendixfixedtable}}{\end{appendixfixedtable}}
\begin{table*}[t]
\centering
\footnotesize
\setlength{\tabcolsep}{2.2pt}
\resizebox{\textwidth}{!}{%
\begin{tabular}{@{}l rrrrr rrrrr rrrrr@{}}
\toprule
 & \multicolumn{5}{c}{DeepSeek-V4.1-Flash} & \multicolumn{5}{c}{Qwen3.8-Flash-Next} & \multicolumn{5}{c}{LongCat-Flash-Lite} \\
\cmidrule(lr){2-6}\cmidrule(lr){7-11}\cmidrule(l){12-16}
Setting & bpp & GB & KL & KL$_{\max}$ & top-1 & bpp & GB & KL & KL$_{\max}$ & top-1 & bpp & GB & KL & KL$_{\max}$ & top-1 \\
\bottomrule
\end{tabular}}
\caption{Full storage and output-fidelity results. KL is the geometric mean over six domains; KL$_{\max}$ is the worst domain; top-1 is next-token agreement with the original. GB counts lookup-table storage. Rates include row/group metadata. QTIP uses within-row rotation and a global scale; CP restores each row norm. RTN uses groups of 64 (80 for Qwen). NVFP4 and MXFP4 include block scales. Bottom: paired KL ratios and 95\% chunk-bootstrap intervals; below 1 favors the first setting.}
\label{tab:main_full}
\end{table*}

\endgroup
\end{minipage}\vspace{12pt}
}]

Table~\ref{tab:main_full} includes the complete rate sweep and two controls that store no table: zeroing every row and replacing each row by the mean of its head.
Removal gives geometric-mean KL values of .773, .308 and 2.89 for DeepSeek, Qwen and LongCat.
A mean row preserves some output behavior in the first two models, but performs worse than removal on LongCat.
The bottom block reports paired ratios: a value below one favors the first configuration.

Table~\ref{tab:ppl} gives WikiText-2 perplexity without benefit normalization.
For example, LongCat changes from 13.316 to 13.538 with the two-bit code, compared with 103.539 after removal.
The large removal gap should not obscure the absolute change caused by compression.

\newpage
\begingroup
\renewenvironment{table}[1][]{\begin{appendixfixedtable}}{\end{appendixfixedtable}}
\begin{table}[t]\centering\small
\begin{tabular}{@{}lrrr@{}}\toprule
Setting & DeepSeek & Qwen & LongCat \\\midrule
Original & 2.611 & 4.671 & 13.316 \\
Table removed & 7.703 & 6.476 & 103.539 \\
RTN 3-bit & 2.835 & 4.757 & 13.681 \\
RTN 2-bit & 3.662 & 5.089 & 14.842 \\
QTIP 2.0 & 3.089 & 4.827 & 13.325 \\
\textbf{CP, 2-bit code} & 3.054 & 4.830 & 13.538 \\
CP, 1.5-bit code & 3.479 & 4.994 & 13.484 \\
\bottomrule\end{tabular}
\caption{WikiText-2 perplexity in its original units (lower is better). Settings match Table~\ref{tab:quality}; no normalization by the table-removal baseline is applied.}
\label{tab:ppl}\end{table}

\endgroup
\clearpage
\section{Source Statistics and Alternative Representations}
\subsection{Row statistics and access counts}
\label{app:stats}

Table~\ref{tab:stats} compares trained rows with dimension-matched Gaussian references.
Read counts come from a sample of 726,000 fineweb-edu documents. 
We tokenize the sample with each model's tokenizer, which gives 727M--755M tokens. 
For each token we compute the rows it reads with our reimplementation of the model's address computation.
The reimplementation uses each model's per-head row offsets and table sizes, and it matches the original addresses bit for bit. 
We then count the reads of every row.
Rows read per token is the total count divided by the number of tokens, which gives the 48, 16 and 12 of Table~\ref{tab:models}. 
These counts set the access skew and the read-weighted statistics of Section~\ref{sec:stats}.
They also set the frequency tiers of Tables~\ref{tab:tools_full} and~\ref{tab:extended_tools} and the hot-row cache of Section~\ref{sec:placement}.
The sample is English web text only.
\begingroup
\renewenvironment{table}[1][]{\begin{appendixfixedtable}}{\end{appendixfixedtable}}
\begin{table}[t]
\centering
\footnotesize
\setlength{\tabcolsep}{2pt}
\resizebox{\columnwidth}{!}{%
\begin{tabular}{@{}lcccc@{}}
\toprule
Statistic & DeepSeek & Qwen & LongCat & Gaussian \\
$d$ & 256 & 160 & 256 & 256/160 \\
\midrule
Kurtosis & 2.99 & 2.94--2.95 & 2.96--2.97 & 2.98/2.96 \\
Kurt.\ (read-wtd.) & 2.98--2.99 & 2.95 & 2.95--2.96 & 2.98/2.96 \\
Max/RMS & 3.00--3.01 & 2.84--2.85 & 2.99 & 3.00/2.86 \\
Cov.\ eff.\ rank & 239--250 & 152--158 & 253--254 & 256/160 \\
KLT gain (dB) & 0.10--0.25 & 0.18--0.53 & 0.04--0.21 & 0 \\
Cols.\ kurt.\,$>$\,5 & 0$^{a}$ & 0 & 0 & 0 \\
\bottomrule
\end{tabular}}
\caption{Six statistics of the hashed tables. Ranges are over layers and n-gram orders (effective rank: over heads); $d$ is the row dimension. Kurtosis and max/RMS (median over rows) are computed within each row (uniformly sampled rows; read-weighted: rows drawn in proportion to their read count). Effective rank is that of the row covariance of each head. The Gaussian column gives the value expected for i.i.d.\ Gaussian rows of the same dimension. KLT gain: coding gain of a covariance KLT at 2 bits, in the value-path readout metric. Cols.: columns with kurtosis above 5, out of 12{,}288, 2{,}560 and 3{,}072 columns. $^{a}$Uniformly sampled rows; with read-weighted sampling, DeepSeek layer-14 order-2 heads have 2{,}048 such columns, an effect of mixing rows of different energy.}
\label{tab:stats}
\end{table}

\endgroup

\paragraph{Normalized coordinates and shared structure.}
Across all three hashed-memory families, normalized within-row kurtosis differs from its Gaussian control by at most .02 for uniform and read-weighted samples, and .06 for the most-read 4,096 rows. 
For 2-gram rows, 16 directions fitted on separate read-heavy rows explain 19\%, 21\%, 33\% and 10\% of the top-4,096 rows' total squared norm in DeepSeek layers 1 and 14, Qwen and LongCat, respectively. 
The dimension-based references are 6.25\% for the 256-dimensional rows and 10\% for Qwen's 160-dimensional rows.
Thus similar within-row statistics coexist with shared directions among frequent rows.

\paragraph{Access frequency and magnitude.}
The top 1\% of rows receive 35\%, 48\% and 33\% of reads in DeepSeek, Qwen and LongCat. 
Mean squared coordinate magnitude correlates with read count positively in Qwen and negatively in DeepSeek; in LongCat the correlation is positive for 2-grams (.53) and near zero for 3- and 4-grams ($-.16$, $-.04$). 
These measurements support separating the row gain from its normalized contents in Section~\ref{sec:stats}.

\subsection{Fitted codes and synthetic rank sweeps}
\label{app:eg-protocol}
\paragraph{Fitted codes on trained tables.}
The row-level comparison evaluates the same frequency-stratified rows used for the statistical analysis, including unread-in-corpus, uniform, read-weighted and top-4,096 sets.
Fitted codebooks use 32,768 disjoint uniform fitting rows for the offline comparison and 200,000 uniform fitting rows for Qwen KL. 
The codebooks fitted by k-means have 256 entries of four values and 65,536 entries of eight values.
Their indices require eight bits per four values and sixteen bits per eight values, respectively, giving the same two-bit code rate. 
Structured variants use the full-table per-head mean and top four PCs.
The byte accounting includes their shared storage and all per-row side information: $16/d$ bpp for a bf16 scale and $64/d$ for four bf16 coefficients. 
The shared codebooks and basis storage are at most 1\,MB per table in these settings. 

For Qwen, we compare the original and compressed models' next-token distributions on the same observed prefixes from 256 FineWeb-Edu documents of 1,024 tokens each.
These documents come from held-out shard 001 and are disjoint from the CPU-side documents of the perturbation experiment below. 
Each looked-up row receives a deterministic coded replacement.
Intervals use 2,000 paired document-bootstrap replicates. 
For unequal rate points, the reported trellis-only comparison is log-linear interpolation across the 1.5-, 2- and 2.5-bit code rates.
The row-level tests cover the three hashed-memory families; output KL for these variants is measured on Qwen.

Across frequency bins and all hashed families, the shared trellis has relative row MSE .0724--.0728, against .098--.099 for the $256\times4$ codebook and .091 for the $65{,}536\times8$ codebook. 
Table~\ref{tab:source-codecs} gives the corresponding Qwen KL comparisons.
The mean-and-PC representation improves on the 2-bit trellis only by using more storage; at its rate, the interpolated trellis-only curve predicts lower KL. This matched-rate value was not measured in a separate run.
Mean subtraction alone gives a smaller improvement of 2.4\% [0.1, 4.5]. 

\begin{table}[t]
\centering
\small
\setlength{\tabcolsep}{3pt}
\begin{tabular}{lrr}
\toprule
Representation & bpp & Qwen KL [95\% CI] \\
\midrule
Shared trellis & 2.1 & .0198 [.0163, .0234] \\
Fitted VQ, $256\times4$ & 2.1 & .0238 [.0193, .0285] \\
Fitted VQ, $65{,}536\times8$ & 2.1 & .0226 [.0184, .0269] \\
Mean + trellis & 2.1 & .0193 [.0160, .0229] \\
Mean + 4 PCs + trellis & 2.5 & .0188 [.0156, .0220] \\
Shared trellis curve & 2.5 & .0144 \\
\bottomrule
\end{tabular}
\caption{Source-code comparisons on held-out Qwen text. Rates include per-row gains and structural coefficients. The final row interpolates the trellis curve in log KL; its uncertainty is not reported. The fitted codebooks and structural variants use the protocol in Appendix~\ref{app:eg-protocol}.}
\label{tab:source-codecs}
\end{table}

\paragraph{Synthetic rank sweep.}
To test whether low rank alone favors fitted codebooks, a separate synthetic rank sweep uses tables of 16,384 rows with $d\in\{160,256\}$ and three seeds. 
A rank-$k$ table has Gaussian latents in a random $k$-dimensional basis, prescribed row norms and a common Haar rotation, stored in bf16. 
The equivalence margin of .05 on mean row kurtosis, mean max/RMS and pooled quantiles was fixed before the run. 
Every basis and codebook is fitted on half of the rows and evaluated on the other half. 
Byte counts include padding to whole bytes and all shared decoder state; each family's best configuration at or below the trellis budget is chosen on the fit rows. 

At $d=256$, rank-$k$ tables pass the Gaussian-coordinate equivalence test for $k\ge16$, yet a PCA-basis code reaches .013 of the trellis MSE for $k=64$ at the 2.06\,bpp budget. 
For spiked full-rank tables with 16 strong directions, the structured-code/trellis MSE ratio is 1.19 when those directions hold a third of the energy, and .94 when they hold half. 
The crossover lies near a spike energy fraction of .45--.5 in this codec grid.
The most-read real 2-gram rows place 10--33\% of their energy in 16 held-out directions (Appendix~\ref{app:stats}), consistent with the trellis advantage in the trained-table comparisons.

\subsection{Other published codes}
Table~\ref{tab:codes} compares the chosen trellis with other published codes.
It compares the E8P lattice code of QuIP\#, AQLM and HIGGS with the trellis code on Qwen in the shared benchmark.
The comparison includes baseline configurations and variants with per-row norm restoration.
With norm restoration on every code, the trellis code has lower KL than E8P, AQLM and HIGGS (paired ratios .877, .875 and .836). 

\begin{table}[t]
\centering
\small
\setlength{\tabcolsep}{3pt}
\resizebox{\columnwidth}{!}{%
\begin{tabular}{@{}l rrrr@{}}
\toprule
Code & bpp & KL & KL$_{\max}$ & top-1 \\
\midrule
  QuIP\# E8P$^{a}$
  & 2.00 
  & .0413 
  & .0605 
  & .934 
  \\
  QuIP\# E8P, norm restored$^{b}$
  & 2.10 
  & .0373 
  & .0518 
  & .937 
  \\
  AQLM $1{\times}16$$^{c}$
  & 2.10 
  & .0385 
  & .0545 
  & .936 
  \\
  AQLM $1{\times}16$, norm restored$^{d}$
  & 2.10 
  & .0374 
  & .0518 
  & .937 
  \\
  HIGGS$^{c}$
  & 2.10 
  & .0404 
  & .0577 
  & .935 
  \\
  HIGGS, norm restored$^{d}$
  & 2.10 
  & .0391 
  & .0544 
  & .936 
  \\
\midrule
  QTIP$^{a}$
  & 2.00 
  & .0347 
  & .0476 
  & .939 
  \\
  CP (trellis, norm restored$^{b}$)
  & 2.10 
  & .0327 
  & .0438 
  & .940 
  \\
\midrule
\multicolumn{5}{@{}l}{\textit{Paired gm-KL ratio [95\% CI]}} \\
  CP / E8P, both norm restored
  & \multicolumn{4}{r}{.877 [.857, .897]} 
  \\
  CP / AQLM, both norm restored
  & \multicolumn{4}{r}{.875 [.853, .897]} 
  \\
  CP / HIGGS, both norm restored
  & \multicolumn{4}{r}{.836 [.819, .856]} 
  \\
  E8P / QTIP
  & \multicolumn{4}{r}{1.192 [1.173, 1.211]} 
  \\
  QTIP / AQLM
  & \multicolumn{4}{r}{.901 [.875, .927]} 
  \\
  QTIP / HIGGS
  & \multicolumn{4}{r}{.858 [.844, .874]} 
  \\
  E8P: norm restored / baseline
  & \multicolumn{4}{r}{.902 [.888, .917]} 
  \\
\bottomrule
\end{tabular}}
\caption{Codes considered for the Engram tables (Section~\ref{sec:recipe}) at about 2 bpp, on Qwen in the shared benchmark of Section~\ref{sec:experiments} (six domains). KL is the geometric mean over domains of the per-token KL to the original model, KL$_{\max}$ the worst domain, top-1 the next-token agreement. $^{a}$Rotation and one scale per table. $^{b}$As in CP: no rotation, one bf16 norm-restoring scale per row. $^{c}$One scale per row (least squares for AQLM, RMS for HIGGS); HIGGS rotates, AQLM does not. $^{d}$The same codes with each row's norm restored. Ratios are paired, with document-chunk bootstrap 95\% CIs.}
\label{tab:codes}
\end{table}

\section{Scale, Rotation, and Allocation}
\label{app:scale}
\subsection{Scale ablation across rates}
\begin{table*}[t]
\centering\footnotesize
\setlength{\tabcolsep}{2pt}
\resizebox{\textwidth}{!}{%
\begin{tabular}{@{}l rrc rrc rrc@{}}
\toprule
 & \multicolumn{3}{c}{DeepSeek} & \multicolumn{3}{c}{Qwen} & \multicolumn{3}{c}{LongCat} \\
\cmidrule(lr){2-4}\cmidrule(lr){5-7}\cmidrule(l){8-10}
Step & bpp & KL & ratio [95\% CI] & bpp & KL & ratio [95\% CI] & bpp & KL & ratio [95\% CI] \\
\midrule
  QTIP
  & 2.00 
  & .0414 
  & -- 
  & 2.00 
  & .0347 
  & -- 
  & 2.00 
  & .0468 
  & -- 
  \\
  No rotation
  & 2.00 
  & .0416 
  & 1.005 {\scriptsize [.995, 1.015]} 
  & 2.00 
  & .0346 
  & .999 {\scriptsize [.974, 1.027]} 
  & 2.00 
  & .0473 
  & 1.010 {\scriptsize [1.000, 1.020]} 
  \\
  Row LS gain
  & 2.06 
  & .0404 
  & .972 {\scriptsize [.964, .981]} 
  & 2.10 
  & .0349 
  & 1.008 {\scriptsize [.984, 1.030]} 
  & 2.06 
  & .0462 
  & .977 {\scriptsize [.974, .981]} 
  \\
  Norm restoration
  & 2.06 
  & .0401 
  & .991 {\scriptsize [.987, .995]} 
  & 2.10 
  & .0327 
  & .937 {\scriptsize [.904, .970]} 
  & 2.06 
  & .0426 
  & .922 {\scriptsize [.917, .928]} 
  \\
\midrule
  CP / QTIP
  & -- & --
  & .968 {\scriptsize [.959, .977]} 
  & -- & --
  & .943 {\scriptsize [.917, .969]} 
  & -- & --
  & .910 {\scriptsize [.900, .919]} 
  \\
  No rotation / rotation, both norm-restored
  & -- & --
  & .999 {\scriptsize [.990, 1.008]} 
  & -- & --
  & .993 {\scriptsize [.971, 1.015]} 
  & -- & --
  & 1.008 {\scriptsize [.997, 1.019]} 
  \\
\midrule
  LS / global, 1.5-bit code
  & 1.56 
  & .0606 
  & .982 {\scriptsize [.975, .991]} 
  & 1.60 
  & .0503 
  & .984 {\scriptsize [.960, 1.010]} 
  & 1.56 
  & .0799 
  & .978 {\scriptsize [.975, .980]} 
  \\
  Norm restoration / LS, 1.5-bit code
  & 1.56 
  & .0593 
  & .979 {\scriptsize [.975, .982]} 
  & 1.60 
  & .0477 
  & .949 {\scriptsize [.925, .967]} 
  & 1.56 
  & .0692 
  & .866 {\scriptsize [.860, .871]} 
  \\
  LS / global, 2.5-bit code
  & 2.56 
  & .0295 
  & .982 {\scriptsize [.973, .990]} 
  & 2.60 
  & .0253 
  & .976 {\scriptsize [.954, .998]} 
  & 2.56 
  & .0275 
  & .985 {\scriptsize [.981, .988]} 
  \\
  Norm restoration / LS, 2.5-bit code
  & 2.56 
  & .0295 
  & 1.000 {\scriptsize [.995, 1.005]} 
  & 2.60 
  & .0250 
  & .987 {\scriptsize [.960, 1.016]} 
  & 2.56 
  & .0264 
  & .959 {\scriptsize [.953, .965]} 
  \\
\bottomrule
\end{tabular}}
\caption{Full ablation from QTIP to CP. The first three step ratios compare each setting with the preceding one: remove within-row rotation at global scale, introduce row-wise LS gain, then restore the norm on the same codes. KL is the geometric mean over six domains. Ratio intervals use paired chunk bootstraps. The lower block repeats gain comparisons at other rates. --: not applicable (the first row has no preceding setting; ratio-only rows have no bpp or KL).}
\label{tab:ablation_full}
\end{table*}

Table~\ref{tab:ablation_full} expands Table~\ref{tab:scale} with rates, absolute KL and paired intervals.
The first comparison removes rotation at a global scale; the next introduces a row-wise LS gain.
The final step changes only the gain on the same codes to restore the original norm.
The lower block repeats the gain comparisons at other rates.

\subsection{Held-out gain test}
\label{app:gain-test}
\paragraph{Scale choices and data.}
We vary only the scale applied to each decoded row, keeping its stored two-bit symbols fixed.
Let $r_i$ be the original row norm, $v_i$ the unit direction of its decoded reconstruction, and $\rho_i$ the cosine between the original and decoded directions, clipped at zero.
We set $\hat e_i(t)=r_i[\rho_i+t(1-\rho_i)]v_i$, so $t=0$ gives LS scaling and $t=1$ restores the original norm.
We test $t\in\{-.5,0,.25,.5,.75,1,1.25,1.5\}$.
All settings store the same symbols and one bf16 scale per row: 2.06, 2.10 and 2.06\,bpp for LongCat, Qwen and DeepSeek.

We calibrate the predictions on 128 FineWeb-Edu documents and test them on 256 different documents from the same held-out shard (001), using 1,024 tokens per document.
We also test 96 chunks of 1,024 tokens each from CodeParrot Python code and Wikipedia in German, Russian, Chinese, Korean and Hindi.
These two sets reuse texts from the six-domain evaluation; they are separate from the web calibration documents, but are not independent samples of the earlier benchmark.
DeepSeek inputs additionally receive a BOS token.

\paragraph{Predictions and evaluation.}
The predictions and decision rules were fixed before each test split was opened, and each test split was evaluated once. 
The primary predictor uses small perturbations of the original table to estimate a local quadratic approximation to output KL; Appendix~\ref{app:gain-analysis} gives the derivation. 
The secondary predictor fits a quadratic to the KL measured on the calibration documents. 
Intervals use 2,000 paired document-bootstrap replicates. 

\begin{table}[t]
\centering\small
\setlength{\tabcolsep}{3pt}
\begin{tabular}{@{}lccc@{}}\toprule
 & Web & Code & Wikipedia \\\midrule
LongCat & .904 & .940 & .904 \\
 & {\scriptsize [.886, .923]} & {\scriptsize [.923, .958]} & {\scriptsize [.889, .918]} \\
Qwen & .983 & .944 & .988 \\
 & {\scriptsize [.975, .992]} & {\scriptsize [.928, .958]} & {\scriptsize [.970, 1.005]} \\
DeepSeek & .997 & .997 & .988 \\
 & {\scriptsize [.987, 1.006]} & {\scriptsize [.979, 1.014]} & {\scriptsize [.971, 1.004]} \\
\bottomrule\end{tabular}
\caption{Norm-restoration KL divided by LS KL in the gain test, with paired 95\% bootstrap intervals. Values below 1 favor norm restoration. Stored codes and storage costs are fixed within each model. Web is the held-out FineWeb-Edu test split; code and Wikipedia reuse the benchmark texts described above.}
\label{tab:gain-heldout}
\end{table}

\paragraph{Observed gains and prediction errors.}
Table~\ref{tab:gain-heldout} reports the comparison between norm restoration and LS on each test set.
Norm restoration lowers KL relative to LS on the test split by ratios of .904--.940 on LongCat and .944--.983 on Qwen web and code; the Qwen multilingual interval and all three DeepSeek intervals include 1. 
The minima of quadratic fits to the test curves lie at $t=.92$--$1.30$ across the nine model--domain combinations, and shrinking below LS ($t=-.5$) raises KL in all three models. 
The primary predictor missed the optimum on LongCat (1.44 against .92--1.05), had the wrong sign on Qwen web and code text, and gave an off-grid value ($-7.0$) on DeepSeek. 
Its finite-difference noise is large relative to the small change from LS to norm restoration, and router choices changed at 40--47\% of token--layer positions, so the local quadratic does not describe the 2-bit operating point. 
The secondary predictor had the right sign in all five cells where the test resolved the effect, and the effect grew under domain shift in two cells. 

\subsection{Rotation, allocation and protected rows}
\label{app:tools}

Tables~\ref{tab:tools} and~\ref{tab:tools_full} report supplementary comparisons for the representation choices in Sections~\ref{sec:tools} and~\ref{sec:recipe}.
Each row is one tool.
It gives the statistical motivation, the measurement on each model and the protocol.
(S) marks the six-domain benchmark described in Appendix~\ref{app:repro}: 262,016 tokens in each of web, code, math, chat and multi5, and 491,280 tokens in multi15. 
(E) marks a separate evaluation with fewer tokens and its own baseline.
Each row states the numerator and denominator of its KL ratio. A value below 1 favors the numerator setting; for example, the rotation comparison uses KL without rotation divided by KL with rotation. 
Per-domain ratios are point values.
They show the pattern of frequency allocation: lower KL on web text and higher KL on code and multilingual text.

\begin{table}[t]
\centering
\footnotesize
\setlength{\tabcolsep}{3pt}
\resizebox{\columnwidth}{!}{%
\begin{tabular}{@{}lcccc@{}}
\toprule
Tool & DeepSeek & Qwen & LongCat & Eval. \\
\midrule
Rotation$^{a}$ & .999 & .993 & 1.008 & S \\
{\scriptsize (Hadamard)} & {\scriptsize [.990, 1.008]} & {\scriptsize [.971, 1.015]} & {\scriptsize [.997, 1.019]} & \\
\addlinespace[2pt]
Second-order & -- & 1.46 & 1.11 & E \\
{\scriptsize rounding} & & {\scriptsize Lloyd--Max} & {\scriptsize GPTQ} & \\
\addlinespace[2pt]
Frequency & -- & 1.044$^{b}$ & -- & S \\
{\scriptsize allocation} & & {\scriptsize [1.016, 1.076]} & & \\
\addlinespace[2pt]
Start-row & .999 & -- & .991 & S \\
{\scriptsize protection} & {\scriptsize [.995, 1.002]} & & {\scriptsize [.988, .993]} & \\
\bottomrule
\end{tabular}}
\caption{Supplementary tool comparisons on hashed tables (frequency allocation lowers KL only on web, the distribution its read counts come from: .915 there, 1.044 over all domains). Entries: KL with / without the tool (S rows: geometric mean over domains, 95\% CI). S: shared benchmark; E: separate evaluation with its own baseline; --: not tested; details in Table~\ref{tab:tools_full}. A pure low-rank replacement was not evaluated end to end (covariance effective rank 239--254 of $d=256$ and 152--158 of $d=160$); a mean + 4 PCs + trellis residual was (Section~\ref{sec:tools}). $^{a}$KL without / with rotation, both with norm restoration, as in Section~\ref{sec:recipe}. $^{b}$Trellis codes in three tiers by read count (top tier 35\% of rows) vs the 2-bit code alone at the same mean rate; the 10\% top tier and per-domain ratios are in Table~\ref{tab:tools_full}.}
\label{tab:tools}
\end{table}

\begin{table*}[t]
\centering
\footnotesize
\resizebox{\textwidth}{!}{%
\begin{tabular}{@{}p{0.12\textwidth}p{0.20\textwidth}p{0.37\textwidth}p{0.20\textwidth}@{}}
\toprule
\raggedright Tool & \raggedright Motivation & \raggedright Measurement & \raggedright Evaluation setup \tabularnewline
\midrule
\raggedright Rotation (random Hadamard within a row) & \raggedright Gaussian-like row statistics motivate testing whether within-row rotation improves reconstruction. & \raggedright KL ratio without/with rotation, both with norm restoration: DeepSeek .999 [.990, 1.008]$^{\dagger}$, Qwen .993 [.971, 1.015], LongCat 1.008 [.997, 1.019]. Table MSE ratio .998--1.001 on all 11 order tables. & \raggedright KL: shared benchmark (six domains; token counts in Appendix~\ref{app:repro}), 2.06--2.10\,bpp. MSE: CPU encode, 2{,}048--4{,}096 read-weighted rows per table. \tabularnewline
\addlinespace
\raggedright Second-order rounding (GPTQ, LDLQ, metric-aware scalar) & \raggedright A readout-only surrogate metric is $W^\top W$: condition number 2.5 (LongCat), metric effective rank 243--254/256 and 155--158/160. This surrogate omits gates and subsequent layers. & \raggedright KL with vs without the metric: LongCat GPTQ .128 vs .115 (vs RTN, 2.5\,bpp); Qwen metric-whitened Lloyd--Max .1502 vs .1027 (2.1\,bpp). Not tested on DeepSeek. & \raggedright (E) Separate evaluation: about 32k web tokens, own baseline per row. \tabularnewline
\addlinespace
\raggedright Low-rank decomposition & \raggedright Covariance effective rank 239--254/256 and 152--158/160: high global rank; frequent-row structure is examined separately in Section~\ref{sec:stats}. & \raggedright Global covariance alone does not determine compression value. A pure low-rank replacement was not evaluated end to end. A mean + 4 PCs + trellis residual was, at equal bytes (Section~\ref{sec:tools}). & \raggedright Qwen KL (Section~\ref{sec:tools}) \tabularnewline
\addlinespace
\raggedright Frequency-based bit allocation & \raggedright Frequently read rows have the same distribution (read-weighted kurtosis), so equal rate gives equal relative error (LongCat: .0737 unread vs .0722 for 10--99 reads). & \raggedright (S) Qwen, KL with / without tiers: trellis codes at 2.6/2.1/1.6\,bpp by read count vs the 2-bit code alone, same mean rate: top tier 35\% of rows (89\% of web reads) 1.044 [1.016, 1.076]; web .915, code 1.074, multilingual 1.092--1.123. With top tier 10\% of rows (72\% of web reads) 1.012 [.980, 1.046]; web .929, code 1.027, multilingual 1.016--1.048. (E) Scalar code, 2.1\,bpp, vs the same code at a uniform rate: web .0853 vs .1030 ($-$17\%), but 1.33--1.44$\times$ higher on code, Chinese and Japanese. & \raggedright (S) Shared benchmark (six domains; token counts in Appendix~\ref{app:repro}), 2.10\,bpp. (E) Separate evaluation (131{,}071 web tokens, plus code, Chinese, Japanese). \tabularnewline
\addlinespace
\raggedright Start-of-sequence row protection & \raggedright Protection tests whether repeatedly accessed start rows warrant additional precision. & \raggedright Rows read at every sequence start kept at original precision (DeepSeek 24 per Engram layer, LongCat 27{,}811). KL ratio vs CP: DeepSeek .999 [.995, 1.002] (random-row control 1.000 [1.000, 1.000]); LongCat .991 [.988, .993] (control .999 [.997, 1.001]). & \raggedright Shared benchmark (six domains; token counts in Appendix~\ref{app:repro}), on top of CP at 2.06\,bpp. \tabularnewline
\bottomrule
\end{tabular}}
\caption{Detail of Table~\ref{tab:tools} (Appendix~\ref{app:scale}). Supplementary tool comparisons on hashed lookup tables (frequency allocation lowers KL only on web, the distribution its read counts come from, and raises it on code and multilingual text). KL ratios are geometric means over the six domains with paired chunk-bootstrap 95\% CIs; per-domain ratios are point values; rotation: KL without / with rotation; frequency allocation and start-row protection: KL with / without the tool. (S): shared benchmark of Section~\ref{sec:setup}; (E): a separate evaluation with fewer tokens and its own baseline, given in the last column. $^{\dagger}$The expected outcome was fixed before this measurement.}
\label{tab:tools_full}
\end{table*}

The extended checks in Table~\ref{tab:extended_tools} further separate average quality from distribution-specific gains.
A top-10\% frequency tier improves DeepSeek's geometric-mean KL by .9\% (ratio .991 [.985, .997]), while multilingual ratios rise to 1.023 and 1.016. 
The top-35\% tier gives ratios 1.010 [1.000, 1.020] on DeepSeek and 1.006 [.995, 1.017] on LongCat; LongCat's code ratio rises to 1.085 [1.057, 1.113]. 
On Qwen, restoring the rows selected at sequence starts improves five non-chat domains (ratios .970--.983, each interval below 1), while the chat estimate is noisy at 1.179 [.997, 1.420]. 
Its six-domain geometric mean is 1.008 [.980, 1.039]. 
These effects make the allocation depend on the deployment distribution.

\begin{table}[t]
\centering\footnotesize
\setlength{\tabcolsep}{2pt}
\resizebox{\columnwidth}{!}{%
\begin{tabular}{@{}lccc@{}}
\toprule
Step & DeepSeek & Qwen & LongCat \\
\midrule
  Top-35\% tiers / uniform & 1.010 & 1.044 & 1.006 \\[-1pt]
  & {\scriptsize [1.000, 1.020]} & {\scriptsize [1.016, 1.076]} & {\scriptsize [.995, 1.017]} \\[2pt]
  Top-10\% tiers / uniform & .991 & 1.012 & -- \\[-1pt]
  & {\scriptsize [.985, .997]} & {\scriptsize [.980, 1.046]} &  \\[2pt]
  Start rows / uniform & -- & 1.008 & -- \\[-1pt]
  &  & {\scriptsize [.980, 1.039]} &  \\[2pt]
  Random rows / uniform & -- & 1.035 & -- \\[-1pt]
  &  & {\scriptsize [.994, 1.087]} &  \\[2pt]
\bottomrule
\end{tabular}}
\caption{Extended allocation checks. Entries are geometric-mean KL ratios to uniform CP with paired 95\% chunk-bootstrap intervals over six domains. Frequency tiers keep the same mean rate; the top-tier fraction refers to rows. Qwen start-row protection restores 37,750 selected rows to bf16 and adds .0019 bpp; the random-row control uses the same count.}
\label{tab:extended_tools}
\end{table}

\section{Geometry and Error Analysis}
\label{app:geometry}

\subsection{Gaussian coordinates with a fixed Gram matrix}

This construction establishes compatibility between Gaussian coordinates and learned geometry, not how training produces them.
Initialization and intermediate checkpoints for the target models are unavailable.
Rotating a fixed table $Z$ by an orthogonal matrix $R$ keeps its Gram matrix, $ZR(ZR)^\top=ZZ^\top$.
If $R$ is Haar-distributed, the normalized direction of any fixed nonzero row is sphere-uniform by rotational invariance.
It can be represented as $h/\lVert h\rVert_2$, with $h\sim\mathcal N(0,I_d)$.
For fixed $k$, $\lVert h\rVert_2/\sqrt d\to1$ in probability, so the first $k$ coordinates scaled by $\sqrt d$ converge in distribution to $\mathcal N(0,I_k)$; see \citet{diaconis1987dozen} for finite-dimensional approximation results.

The same rotation applies to every row, preserving all learned pairwise geometry and rank.
Rows can remain correlated, including a rank-one table, and a unit vector's full coordinates obey a fixed-norm constraint.
The construction is a compatibility argument for source statistics.
Actual trained coordinates require the empirical checks in Section~\ref{sec:stats}; it does not model the optimizer or establish that the complete network is invariant to a coordinate change with the consumer held fixed.

\subsection{Fixed-shape gain and local output distortion}
\label{app:gain-analysis}
With $e=ru$, $\lVert u\rVert=\lVert v\rVert=1$ and $c=u^\top v\in(0,1)$,
\begin{align}
\lVert gv-ru\rVert^2 &= (g-rc)^2+r^2(1-c^2),\\
D(r)-D(rc) &= r^2(1-c)^2.
\end{align}
Thus $rc$ minimizes coordinate MSE, while $r$ preserves magnitude.
For a local quadratic metric $H\succeq0$ of the downstream network, holding the other rows fixed, the distortion is
\begin{equation}
\mathcal D_H(g)=\tfrac12(gv-ru)^\top H(gv-ru).
\end{equation}
Writing $A=v^\top Hv>0$ and $B=v^\top Hu$, differentiating the quadratic distortion gives $g_H^*=rB/A$.
The difference between norm restoration and LS is
\begin{equation}
\mathcal D_H(r)-\mathcal D_H(rc)
=\tfrac12r^2(1-c)\big[(1+c)A-2B\big].
\end{equation}
For $c\in(0,1)$, norm restoration beats LS in this metric exactly when
\begin{equation}
2v^\top Hu>(1+c)v^\top Hv.
\label{eq:norm-condition}
\end{equation}
For $H=I$, LS is optimal.
Equation~\ref{eq:norm-condition} identifies when preserving magnitude is preferable despite the extra coordinate error.
This is an unconstrained fixed-shape result.
If the gain is restricted, its optimum is projected onto the allowed interval.

For output KL, a local metric at the original table can be written $H=\mathbb E_x[J_x^\top \mathcal F_x J_x]$, where $J_x$ differentiates model logits with respect to the table and $\mathcal F_x=\operatorname{diag}(p_x)-p_xp_x^\top$ is the categorical Fisher matrix.
The expectation includes selective and repeated row use.
A single-row block treats the other rows as fixed; simultaneous table changes also involve cross-row terms.
The readout surrogate $W_v^\top W_v$ omits the gate and subsequent network.

For a fixed-code gain sweep, define
\begin{equation}
\hat e_i(t)=r_i[\rho_i+t(1-\rho_i)]v_i,
\end{equation}
where $\rho_i=u_i^\top v_i$ is the row cosine.
Stacking all row errors gives $\epsilon(t)=\epsilon_0+t s$.
The local approximation is
\begin{equation}
\KL(t)\approx\tfrac12\epsilon_0^\top H\epsilon_0+a t+\tfrac12b t^2,
\end{equation}
with $a=s^\top H\epsilon_0$ and $b=s^\top Hs$.
For $b>0$, it predicts $t^*=-a/b$ and a norm-restoration--LS difference of $a+b/2$.
Appendix~\ref{app:gain-test} tests this prediction on held-out text.
Large distortion, expert-routing changes and a poorly estimated metric can invalidate the local approximation.

\subsection{Perturbations through the readout}
A separate perturbation experiment adds to every looked-up Qwen row an error with squared norm $0.05\lVert e\rVert^2$ or $0.2\lVert e\rVert^2$. 
The error lies along a chosen subspace: norm only, direction only, isotropic, or the top or bottom 16 right-singular directions of each head's value or key projection. 
A row has the same error wherever it is read. 
We compare next-token distributions on identical observed prefixes from 256 held-out FineWeb-Edu sequences of 1,024 tokens. These come from shard 001, excluding the first 400 documents used for the CPU readout analysis; KL ratios are paired on the same tokens. 

Qwen's raw rows have kurtosis 2.94, while the value outputs $W_v e$ have kurtosis 93.5. 
At error energy $0.2\lVert e\rVert^2$, perturbations along the top 16 right-singular directions of the value projection change the memory readout 1.6 times as much as perturbations along the bottom 16, yet give slightly lower KL (ratio .92 [.88, .96]). 
Pure norm errors incur more KL than equal-energy direction errors (direction/norm .86 [.83, .88]). 

\subsection{LongCat damage model}
\label{app:damage}

\begin{figure}[t]
\centering
\begin{tikzpicture}
\begin{axis}[
  width=0.96\columnwidth, height=6cm,
  xmode=log, ymode=log,
  xmin=5e-6, xmax=0.6, ymin=1.5e-3, ymax=0.3,
  xlabel={Direction error $S$},
  ylabel={KL per token},
  label style={font=\footnotesize}, tick label style={font=\scriptsize},
  legend style={font=\scriptsize, draw=none, fill=white, fill opacity=0.85, text opacity=1, at={(0.02,0.98)}, anchor=north west},
  legend cell align=left,
  ymajorgrids, xmajorgrids, grid style={gray!20},
]
\addplot[gray, dashed, forget plot] coordinates {(5e-6,0.00210202) (0.6,0.00210202)};
\node[font=\scriptsize, gray!80!black, anchor=south east] at (axis cs:0.3,0.00210202) {offset $c_0=.0021$};
\addplot[color={rgb,255:red,0;green,158;blue,115}, line width=0.9pt, no marks] table[col sep=comma, x=S, y=KL_dir] {figures/data/damage_curve.csv};
\addlegendentry{model, norm kept ($dN=0$)}
\addplot[color={rgb,255:red,213;green,94;blue,0}, line width=0.9pt, densely dashed, no marks] table[col sep=comma, x=S, y=KL_mmse] {figures/data/damage_curve.csv};
\addlegendentry{model, MSE scale ($dN=\sqrt{1-S}-1$)}
\addplot[only marks, mark=square*, mark size=1.7pt, color={rgb,255:red,213;green,94;blue,0}, fill opacity=0.8] table[col sep=comma, x=S, y=KL] {figures/data/damage_mmse.csv};
\addlegendentry{MSE-scaled codecs (fit)}
\addplot[only marks, mark=*, mark size=1.9pt, color={rgb,255:red,0;green,158;blue,115}] table[col sep=comma, x=S, y=KL] {figures/data/damage_ncc.csv};
\addlegendentry{norm-restored codecs (fit)}
\addplot[only marks, mark=triangle*, mark size=2.6pt, color={rgb,255:red,0;green,114;blue,178}] table[col sep=comma, x=S, y=KL] {figures/data/damage_dir.csv};
\addlegendentry{pure direction edits (held out)}
\end{axis}
\end{tikzpicture}
\caption{LongCat damage model $\mathrm{KL}=c_0+k\,S^{p}+k_{\parallel}\,dN^2$ with the measured small-perturbation offset
$c_0=.0021$ and $k=.198$, $p=.79$,
$k_{\parallel}=2.19$ fitted once on 29 codec configurations
(LM, VQ and TCQ at 0.5--4 bits with the MSE-optimal scale or with norm restoration, plus one global gain, not drawn).
Pure direction edits that keep every row norm (triangles) are out of sample. $S$ and $dN$ summarize direction and norm error using the codec-level approximations defined in the text; the direction controls set $S$ explicitly.}
\label{fig:damage}
\end{figure}
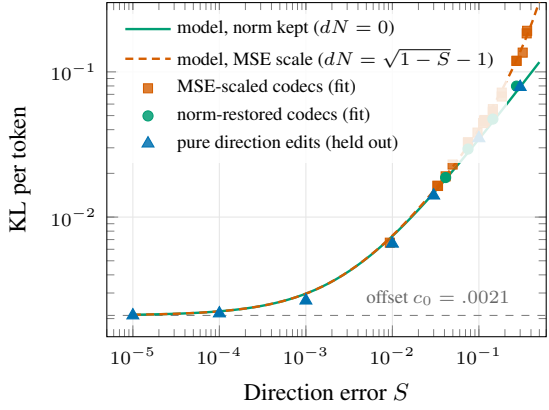

The damage model summarizes codec error with a direction-error proxy $S$ and a relative norm-change proxy $dN$.
For each codec, $D$ is the arithmetic mean over LongCat's twelve tables of $\sum_i\|\hat e_i-e_i\|^2/\sum_i\|e_i\|^2$.
For LS-scaled codes we use $S=D$ and $dN=\sqrt{1-D}-1$; norm-restored variants retain the corresponding LS value of $S$ and set $dN=0$.
For the norm-expanding family, we use $S=D/(1+D)$ and $dN=\sqrt{1+D}-1$; a uniform multiplicative gain $1+g$ gives $S=0$ and $dN=g$.
These are codec-level approximations, not direct averages of measured row angles and norm changes.
The pure-direction controls set $S$ explicitly and preserve each row's norm.
Output KL is evaluated on 32,768 FineWeb-Edu tokens, arranged as sixteen 2,048-token sequences, from the stream after its first 200,000 documents.
The model is
\begin{equation}
\mathrm{KL} = c_0 + k\,S^{p} + k_{\parallel}\,dN^2 .
\end{equation}
The offset $c_0=.0021$ is the KL measured after a very small pure-direction perturbation ($S=10^{-5}$) on LongCat.
It represents a measured perturbation response, rather than a universal lower bound on KL; an unperturbed repeat in the same process gives zero KL. 
The other constants are $k=.198$, $p=.79$ and $k_{\parallel}=2.19$. 
We fit them once on 29 LongCat codec configurations, with a log-RMSE of .053. 
Figure~\ref{fig:damage} plots KL against $S$ for these configurations.
It provides a descriptive comparison with the fixed-code gain tests.

The model matches the codecs it was fitted on to within about 10\%. 
It also predicts the held-out pure direction edits, the triangles in the figure.
For $S$ from $10^{-5}$ to .3, measured KL is 0.89--1.01 times the prediction. 
The model fails in three cases.
It overpredicts the gain of norm restoration at 1.5 bits about twofold. 
It underpredicts errors that inflate the norm at relative MSE .13--.3 by 1.3--2.0$\times$. 
It also misses a coherent gain of $\pm$3\% on all rows by 1.5--2.0$\times$. 
We fitted and checked the model on LongCat only, so we use it as a description.

\section{Storage Format and Systems Measurements}
\label{app:serving}

End-to-end serving runs and the unfused lookup controls use one node with 8 H200 GPUs, two Xeon Platinum 8462Y+ CPUs, 2\,TB of DRAM and 6 local PM1733a NVMe drives in RAID-0. 
Each H200 has 141\,GB of HBM nominally; nvidia-smi reports 143,771\,MiB, or 150.8 $\times10^9$ bytes. 
Peak GPU memory in Section~\ref{sec:placement} and Table~\ref{tab:placement_full} is PyTorch's \texttt{max\_memory\_allocated} in units of $10^9$ bytes, taken on the GPU with the largest value. 
It counts allocated tensors only, not memory the allocator has reserved but not handed out.

\begin{table*}[t]
\centering
\small
\resizebox{\textwidth}{!}{%
\begin{tabular}{@{}p{1.9cm}p{2.9cm}p{5.9cm}p{3.7cm}@{}}
\toprule
Tier & Who needs it & What 2\,bpp changes (measured) & What it does not change \\
\midrule
\raggedright GPU HBM, 4 GPUs &
\raggedright RL rollouts and public inference code that keep the table on the GPU &
\raggedright FP8 needs 50.7\,GB/GPU 
and runs out of memory at decode batch 512 
and at 64k-token prefill; 
in the representative setting its prefill takes 64.66 $\pm$ 2.49\,s (peak 147.0\,GB). 
NVFP4 (27.6\,GB/GPU) runs both, but its 64k-token prefill step takes 59.81 $\pm$ 4.75\,s. 
CP needs 12.7\,GB/GPU, 
runs every workload, and prefills in 13.27 $\pm$ 0.22\,s (peak 109.5\,GB) and 64k tokens in 18.23 $\pm$ 0.01\,s. 
&
\raggedright Decode step time (795.7 vs.\ 792.9\,ms, representative setting) and prefill steps up to 16k tokens (within 0.9\%). Prefill slows when a large table leaves little HBM (left). 
\tabularnewline
\addlinespace
\raggedright GPU HBM, 8 GPUs &
\raggedright Same, with the model spread over a full node &
\raggedright Both fit (25.3 vs.\ 6.3\,GB/GPU); 
CP leaves 19.0\,GB/GPU more for KV cache and raises the per-step ceiling from 
decode batch 1024 to $\ge$2048 
and from 96k to $\ge$128k prefill tokens. 
&
\raggedright Step time (within 2.4\% at every point both ran). 
\tabularnewline
\addlinespace
\raggedright Host DRAM &
\raggedright Several model replicas per node; hosts with little RAM &
\raggedright Table size: 202.7 $\to$ 50.7\,GB; 
Peak node RSS 225 $\to$ 75\,GB. 
Lookup alone at 64k tokens: 53.5 $\to$ 19.4\,ms. 
Smallest host that holds the table (calc.): 256 GB $\to$ 64 GB. 
&
\raggedright End-to-end time: the lookup is $\le$0.8\% of a forward step. 
\tabularnewline
\addlinespace
\raggedright SSD &
\raggedright DRAM smaller than the table (long tail on SSD) &
\raggedright Nothing per read: each row is one 4\,KB read in both formats, 
48 reads per decoded token. 
&
\raggedright Lookup latency; with prefetch the exposed lookup at 16k-token prefill is 2.7\,ms (FP8) 
and 1.5\,ms (CP), vs.\ 199.2\,ms without. 
\tabularnewline
\addlinespace
\raggedright Hot-row cache (HBM) + SSD &
\raggedright Same, with spare HBM &
\raggedright At 12.9\,GB the cache serves 50.6\% $\to$ 70.7\% of reads; 
a 51.5\,GB cache holds the whole 50.7\,GB CP table and serves 100\% of reads (FP8 rows: 70.7\%). 
Simulated SSD reads/token at 32\,GB: 19.4 $\to$ 2.7. 
&
\raggedright Decode still misses every step (FP8 with 12.9\,GB cache: 20.8 SSD reads/token at batch 64), 
so the off-GPU round trip stays on the critical path.
\tabularnewline
\bottomrule
\end{tabular}}
\caption{Placement summary (detail for Section~\ref{sec:placement}): what compressing the DeepSeek tables to 2\,bpp changes in each memory tier, and what it does not (8$\times$H200 node,
reference inference code; representative setting: batch 8, 4k-token prompt, 256 decode steps). 4-GPU timings are mean $\pm$ s.d.\ over 3 runs; other settings are one run each.
Peak: maximum over GPUs of PyTorch's allocated memory, in $10^9$ bytes (an H200 has 141\,GB nominal).
``Calc.'' marks a calculation (table only, 87.5\% of installed RAM usable); everything else is measured.}
\label{tab:placement_full}
\end{table*}

\subsection{Records and lookup kernels}
\label{app:format}

Each CP row is a fixed-size record.
The record holds the trellis bitstream of the row and one bf16 scale $\alpha_i$ (Section~\ref{sec:recipe}).
A $b$-bit code on a $d$-value row takes $bd/8$ bytes, and the scale adds 2 bytes. 
The rate is therefore $b + 16/d$\,bpp. 
For DeepSeek at $b=2$ and $d=256$, a record is 64 + 2 = 66\,B, or 2.06\,bpp. 
For Qwen at $d=160$, it is 40 + 2 = 42\,B, or 2.10\,bpp. 
On SSD the records are aligned so that no record crosses a 4\,KB page, and one row is one read. 

\begin{table}[t]
\centering
\footnotesize
\setlength{\tabcolsep}{2.5pt}
\resizebox{\columnwidth}{!}{%
\begin{tabular}{@{}lrrrr@{}}
\toprule
 & & \multicolumn{2}{c}{HBM (ms)} & DRAM (ms) \\
\cmidrule(lr){3-4}\cmidrule(l){5-5}
Layout & KB/token & decode & prefill & prefill \\
\midrule
  Rows (CP)
  & 3.2 
  & 0.0357 
  & 0.228 
  & 6.70 
  \\
  Tiles, whole tile
  & 50.7 ($16{\times}$) 
  & 0.133 
  & 3.64 ($16{\times}$) 
  & 9.90 ($1.48{\times}$) 
  \\
  + Hadamard, 256 rows
  & 811 ($256{\times}$) 
  & 1.66 
  & 101 ($443{\times}$) 
  & 187 ($28{\times}$) 
  \\
\midrule
\multicolumn{5}{@{}l}{\textit{Fused gather and decode: mean of run medians}} \\
Rows (CP) & 3.2 & 0.0252 & 0.19 & 6.54 \\
Tiles, row windows & 7.3 & 0.0255 & 0.221 & 10.8 \\
\bottomrule
\end{tabular}}
\caption{DeepSeek lookup costs at two code bits (48 rows/token; one H200). HBM: GPU memory; DRAM: pinned host memory over PCIe. Decode: batch 64; prefill: 4,096 tokens. Top: separate gather/decode, single-session medians; parentheses give factors over row records. Bottom: mean of five independent run medians (100 calls after 20 warm-ups each); row windows use random payloads, CP uses model codes. Blocks use different nodes and protocols and should be compared only within each block. KB/token is computed code traffic, not measured bandwidth; this test does not compare output KL.}
\label{tab:layout}
\end{table}

Table~\ref{tab:layout} compares lookup layouts on DeepSeek.
The fused comparison uses one H200 on a four-GPU node with Xeon Platinum 8462Y+ CPUs, PyTorch 2.14.1+cu130 and Triton 3.8.0.
CPU execution and host allocations are bound to NUMA node 0; GPU clocks are not locked.
The full 768,022,850-row tables use real CP codes and random tile payloads, with the same real row scales and decoder lookup table.
Each kernel passes bit-exact checks against its own reference decoder on 8,192 rows, including table boundaries; the two layouts are not matched for output KL.

Lookup requests come from the DeepSeek-tokenized first 30,000 documents of held-out FineWeb-Edu shard 001.
We test prefill lengths of 1,024, 4,096 and 16,384 tokens and decode batches of 1 and 64, using five independent processes per placement.
Each workload has 20 warm-up requests and 100 timed requests per method; identical requests are paired across methods with alternating execution order.
Different runs use different request seeds, so between-run variation includes both execution and input variation.
Timing uses a host clock with GPU synchronization before and after each call, including output allocation but excluding data loading and request transfer.
No cache flush is used.
We report the mean and sample standard deviation of the five run medians, not statistics treating all 500 calls as independent runs.

At 4,096 tokens, CP and tile windows take $0.1905\pm0.0034$ and $0.2210\pm0.0034$\,ms in HBM, and $6.544\pm0.022$ and $10.781\pm0.091$\,ms from host DRAM.
The per-run tile-window/CP time ratios average $1.160\pm0.003$ and $1.648\pm0.017$, respectively.
Randomizing the CP code payload while preserving scales gives ratios of $1.163\pm0.003$ and $1.674\pm0.031$ under the same protocol, retaining the direction and approximate size of the prefill advantage.
All 20 processes across both payload conditions and placements pass correctness checks, with 20,000 timed calls in total.
HBM decode differences are small, and the batch-64 ordering reverses in one of five runs with real CP codes; we do not claim a consistent HBM decode advantage.

The unfused controls in the top block use separate gather and decode kernels on the eight-GPU node described above; compare layouts within each block.
The row-window decoder needs 7.3\,KB of window bits per token, spanning 24.6\,KB of 32-byte sectors, 5.3$\times$ the sectors of row records.
These byte counts are computed from the layout, since hardware memory counters were unavailable.

A separate CP--FP8 benchmark on the eight-GPU node uses a Triton kernel that reads and decodes each row in one pass.
With the DeepSeek table in HBM, lookup takes 0.04\,ms for one decode token and 0.83\,ms for a 16k-token prefill. 
Reading the FP8 table takes 0.075\,ms and 2.45\,ms, measured on a quarter shard that fits on one GPU. 

\subsection{Memory tiers and workload measurements}

SSD reads pass through host memory because GPUDirect Storage was unavailable on the test system.

Table~\ref{tab:tiers} lists the memory tiers of this node, as summarized in Section~\ref{sec:bg-tiers}.
It gives bandwidth, read unit, latency and lookup time for each tier.
We measured the SSD with fio 3.38 using io\_uring, O\_DIRECT and 4\,KB random reads. 
At queue depth 1 it gives 10.8k IOPS and 91\,$\mu$s per read. 
With 32 jobs at queue depth 64 it gives 2.56M IOPS, and 64\,KB reads reach 40\,GB/s. 
O\_DIRECT reads below 4\,KB are not possible on this path, so every format needs one 4\,KB read per row. 

\begin{table*}[t]
\centering
\footnotesize
\resizebox{\textwidth}{!}{%
\begin{tabular}{@{}p{0.06\textwidth}p{0.085\textwidth}p{0.165\textwidth}p{0.065\textwidth}p{0.095\textwidth}p{0.09\textwidth}p{0.27\textwidth}@{}}
\toprule
\raggedright Tier & \raggedright Capacity (node) & \raggedright Bandwidth (measured) & \raggedright Read unit & \raggedright Latency & \raggedright Lookup, 64k tok (ms) FP8 $\to$ CP & \raggedright What compression changes \tabularnewline
\midrule
\raggedright GPU HBM & \raggedright 1{,}128 GB (8$\times$H200) & \raggedright 4{,}579 GB/s read per GPU & \raggedright $\le$64~B$^{a}$ & \raggedright -- & \raggedright 9.70 $\to$ 3.23$^{b}$ & \raggedright Whether the table fits next to the model weights. \tabularnewline
\raggedright DRAM & \raggedright 2 TB & \raggedright STREAM triad 191--315 GB/s$^{c}$; host$\to$GPU 55.2 GB/s & \raggedright 64~B$^{a}$ & \raggedright 1.16 $\mu$s (PCIe round trip) & \raggedright 53.5 $\to$ 19.4 & \raggedright Host RAM per replica (202.7 $\to$ 50.7 GB) and PCIe traffic (12.7 $\to$ 3.2 KB per token). \tabularnewline
\raggedright Local SSD & \raggedright 6 NVMe, RAID-0 & \raggedright 4 KB random 2.56M IOPS; 64 KB reads 40 GB/s & \raggedright 4 KB & \raggedright 91 $\mu$s (QD1) & \raggedright 1{,}243 $\to$ 1{,}048 & \raggedright Nothing on the read path: one 4 KB read per row in every format (16k-token prefill: reads 232 vs 228 ms). The gap is CPU-side row extraction (15.2 vs 1.0 ms). More rows fit in the same cache. \tabularnewline
\bottomrule
\end{tabular}}
\caption{Memory tiers (detail for Section~\ref{sec:placement}): one 8$\times$H200 node and the effect of compressing the DeepSeek Engram tables from FP8 (264 B/row) to CP at 2.06\,bpp format (66 B/row). Lookup is the median time to fetch the 48 rows of each of 64k tokens on one GPU. Bandwidths: nvbandwidth (HBM, PCIe) and STREAM (DRAM); fio (SSD). $^{a}$Access granularity of the memory system, not measured. $^{b}$FP8 uses a quarter shard of the table, since the full FP8 table does not fit one GPU. $^{c}$One NUMA node to both nodes; measured while other jobs used host memory, so a lower bound.}
\label{tab:tiers}
\end{table*}

Table~\ref{tab:placement_full} summarizes effects across memory tiers, complementing the GPU measurements in Table~\ref{tab:capacity}.
It lists for each tier who needs it, what 2\,bpp changes and what it leaves unchanged. 
On 4 GPUs it separates decode, where step time does not change, from prefill, which slows when a large table leaves little HBM for activations.

\subsection{Caching and host placement}
\label{app:host-cache}\label{sec:host-cache}

In the DeepSeek host-memory tests, compressing the table from 202.7 to 50.7\,GB saves capacity and speeds up isolated lookups.
Lookup accounts for less than 1\% of a forward step in our eager implementation without graph capture, so complete step time changes little; the backbone remains on the GPU.

\begin{figure}[t]
\centering
\begin{tikzpicture}
\begin{groupplot}[
  group style={group size=2 by 1, horizontal sep=1.15cm},
  width=0.51\columnwidth, height=4.6cm,
  xmode=log, log basis x=2,
  tick label style={font=\scriptsize}, label style={font=\footnotesize},
  title style={font=\footnotesize, yshift=-1ex},
  ymajorgrids, grid style={gray!20},
  legend style={font=\scriptsize, draw=none, fill=white, fill opacity=0.8, text opacity=1},
  legend cell align=left,
]
\nextgroupplot[title={(a) HBM cache hit rate},
  xlabel={Cache size (GB)}, ylabel={Reads served (\%)},
  xmin=0.8, xmax=64, ymin=20, ymax=105,
  xtick={1,4,16,64}, xticklabels={1,4,16,64},
  legend pos=north west]
\addplot[color={rgb,255:red,0;green,114;blue,178}, mark=*, mark size=1.4pt, line width=0.8pt]
  table[col sep=comma, x=GB, y=hit] {figures/data/cache_hit_fp8.csv};
\addlegendentry{FP8 rows}
\addplot[color={rgb,255:red,213;green,94;blue,0}, mark=square*, mark size=1.3pt, line width=0.8pt]
  table[col sep=comma, x=GB, y=hit] {figures/data/cache_hit_qtip2ncc.csv};
\addlegendentry{2-bpp rows}
\addplot[only marks, mark=o, mark size=3.2pt, black, forget plot,
  nodes near coords, point meta=explicit symbolic,
  nodes near coords style={font=\scriptsize, anchor=east, xshift=-2pt}]
  table[col sep=comma, x=GB, y=hit, meta=label] {figures/data/cache_hit_mark.csv};
\nextgroupplot[title={(b) SSD reads per token},
  xlabel={Cache size (GB)}, ylabel={4\,KB reads / token},
  xmin=0.2, xmax=300, ymin=0, ymax=48,
  xtick={0.25,1,4,16,64,256}, xticklabels={.25,1,4,16,64,256},
  ]
\addplot[color={rgb,255:red,0;green,114;blue,178}, mark=*, mark size=1.4pt, line width=0.8pt]
  table[col sep=comma, x=GB, y=reads] {figures/data/cache_ssd_fp8.csv};
\addplot[color={rgb,255:red,213;green,94;blue,0}, mark=square*, mark size=1.3pt, line width=0.8pt]
  table[col sep=comma, x=GB, y=reads] {figures/data/cache_ssd_b21.csv};
\end{groupplot}
\end{tikzpicture}
\caption{DeepSeek tables behind a static hot-row cache (rows ranked by read count, fineweb held-out reads).
(a) Share of reads served by a cache of a given size holding FP8 or 2-bpp rows (CP, 2.06\,bpp); circles mark the
12.9\,GB/node cache of the end-to-end runs. (b) Simulated SSD reads per token (one 4\,KB read per missed row,
48 rows per token) for the FP8 table (202.7\,GB) and the CP table (2.06\,bpp, 50.7\,GB); colors as in (a).}
\label{fig:cache}
\end{figure}
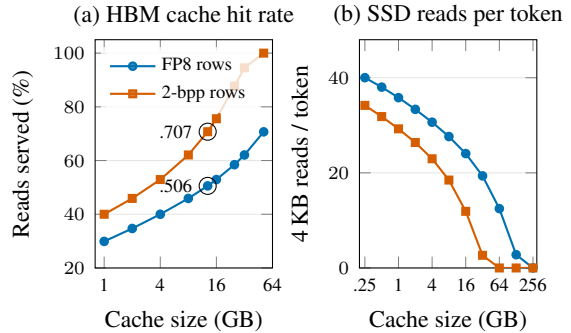

Figure~\ref{fig:cache} comes from a simulation of a static hot-row cache.
The cache holds the rows with the highest FineWeb-Edu read counts (Appendix~\ref{app:stats}), and we evaluate it on held-out FineWeb-Edu reads.
Each missed row costs one 4\,KB read. 
The simulation uses the fio constants above, 2.56M IOPS and 91\,$\mu$s at queue depth 1. 
These constants set only the throughput limits, and the hit rates do not depend on them.
At a cache capacity of 12.9\,GB, CP serves 70.7\% of held-out reads from the cache, compared with 50.6\% for FP8. 

Compression improves the hit rate by fitting more rows in the same cache capacity, while the 4\,KB minimum read unit leaves the cost of an individual row miss unchanged.
A decode step requests many rows, and the tested caches still require off-GPU reads at each step.
Higher hit rates therefore reduce simulated traffic without necessarily eliminating the wait for those reads.

On Qwen (2 H200 GPUs, Hugging Face eager inference), keeping the 102.4\,GB bf16 table in pinned host DRAM costs little time. 
Decoding one token takes 83.7\,ms, against 79.3\,ms with a 13.8\,GB trellis-coded table on GPU (2.15\,bpp, a trellis with a 10-bit state and a stored lookup table). 
At 2.1\,bpp, CP would shrink this table to 13.4\,GB. 

\section{Token-Indexed Tables (Gemma PLE)}
\label{sec:boundary}
\label{app:ple}

Gemma's per-layer embedding (PLE) assigns one row to each token and multiplies the retrieved vector elementwise by a gate.
This changes both the accessed rows and their consumer, providing a boundary test for the results on hashed memories.

\paragraph{Sensitive rows.}
Most PLE rows have per-row kurtosis 3.3--3.5, moderately above the Gaussian value of 3. 
The BOS row and the rows of about 100 frequent tokens have heavy tails. 
These rows are read in a few sensitive layers, and they dominate the loss.
At 2.0\,bpp without rotation, QTIP gives a worst-domain KL of .106 on E2B and .162 on E4B (Table~\ref{tab:ple_rows}). 
Keeping the BOS row and the top-100 rows in bf16 costs 0.006\,bpp and lowers these to .027 and .028. 
This gives similar worst-domain KL to QTIP with rotation (.033 and .029). 
On E2B-it, rotation is better (.126 against .140). 
These token-indexed tables therefore require a different treatment of sensitive rows from the uniform, unrotated representation in Section~\ref{sec:recipe}.
In the tested hashed tables, protecting sequence-start rows changes aggregate KL by at most 1\% on DeepSeek and LongCat (Table~\ref{tab:tools} in Appendix~\ref{app:tools}). 

\begin{table}[t]
\centering
\small
\setlength{\tabcolsep}{3.5pt}
\resizebox{\columnwidth}{!}{%
\begin{tabular}{@{}lccccc@{}}
\toprule
 & \multicolumn{3}{c}{No rotation} & \multicolumn{2}{c}{Rotation} \\
\cmidrule(lr){2-4}\cmidrule(l){5-6}
Model & QTIP & +BOS & +top-100 & QTIP & +top-100 \\
\midrule
E2B
  & .106 
  & .031 
  & \textbf{.027} 
  & .033 
  & .026 
  \\
E2B-it
  & .227 
  & .198 
  & .140 
  & \textbf{.126} 
  & .123 
  \\
E4B
  & .162 
  & .030 
  & \textbf{.028} 
  & .029 
  & .026 
  \\
E4B-it
  & .067 
  & .067 
  & .061 
  & \textbf{.059} 
  & .056 
  \\
3n-E4B
  & .108 
  & .109 
  & \textbf{.105} 
  & .123 
  & .116 
  \\
\bottomrule
\end{tabular}}
\caption{Worst-domain KL of Gemma PLE tables at 2.0\,bpp (six domains) when the BOS row, or the BOS row and the
100 most-read token rows (+top-100), are kept in bf16 in every layer (+0.006\,bpp), without and with rotation. Models: Gemma-4 E2B, E4B and their instruction-tuned (-it) versions, gemma-3n-E4B.
Bold: lowest of the first four columns.}
\label{tab:ple_rows}
\end{table}

\paragraph{Predicting model sensitivity.}
Table statistics alone did not predict this effect.
Before measuring E4B-it, we fixed a prediction of the rotation KL ratio from per-layer kurtosis.
The prediction was .21, and the measurement was .877. 
The table-side part was right: the MSE ratio without and with rotation was 1.74 predicted and 1.78 measured. 
The step from MSE to KL failed.
When only layer 10 is quantized without rotation, KL is .121 on E4B and .004 on E4B-it. 
The two layers have similar statistics, so the gap lies in model sensitivity.
We found this explanation after the prediction missed, so it is post hoc.
A second prediction, also fixed before the measurement, gave the correct decision for both gemma-3n models and the correct magnitude for neither. 

\paragraph{Layer-level norm adjustment.}
On PLE, a layer-mean norm adjustment on top of QTIP's global scale changes worst-domain KL in both directions. 
It raises it by 6.3\% and 9.0\% on E2B-it and by 6.0\% on E4B without rotation. 
In the other five settings (E2B, E4B with rotation, gemma-3n) it lowers it by 3.9--79.8\%. 
This adjustment differs from the per-row norm restoration of CP.

\end{document}